\documentclass[letterpaper]{article}
\usepackage{aaai2026}
\usepackage{times}
\usepackage{helvet}
\usepackage{courier}
\usepackage[hyphens]{url}
\usepackage{graphicx}
\usepackage{natbib}
\usepackage{caption}
\nocopyright

\usepackage{amsmath,amssymb}
\usepackage{booktabs}
\usepackage{array}
\usepackage{multirow}
\usepackage{algorithm}
\usepackage{algpseudocode}

\newcommand{\E}{\mathbb{E}}
\newcommand{\Prob}{\mathbb{P}}
\newcommand{\Var}{\mathrm{Var}}
\newcommand{\cvar}{\mathrm{CVaR}}
\newcommand{\phat}{\hat{p}}

\title{When Agent Automation Becomes Profitable:
Quantifying and Insuring Autonomous AI Risk through Trace-Economic Underwriting}

\author{
    Binyan Xu\textsuperscript{\rm 1},
    Xilin Dai\textsuperscript{\rm 2},
    Fan Yang\textsuperscript{\rm 1},
    Kehuan Zhang\textsuperscript{\rm 1}
}
\affiliations{
    \textsuperscript{\rm 1}The Chinese University of Hong Kong, Hong Kong, Hong Kong\\
    \textsuperscript{\rm 2}Zhejiang University, Hangzhou, China\\
    \{binyxu, yf020, khzhang\}@ie.cuhk.edu.hk, xilin2023@zju.edu.cn
}

\begin{document}
\maketitle


\begin{abstract}
AI agents can now take irreversible actions in operational systems, but agent-caused losses are still not clearly assigned, priced, or transferred. Providers often disclaim consequential damages, users are left with uncompensated losses, and default human review limits the efficiency gains of automation. We ask when autonomous AI deployment can become economically acceptable despite failure risk. Our answer is to quantify risk at the \emph{customer-task-trace} episode level and transfer it through insurance. Automation is acceptable when its expected benefit exceeds the premium, control cost, and remaining risk. This requires a \emph{defined role} with bounded permissions and comparable traces. We introduce \emph{trace-economic underwriting}, which maps tool-use traces to customer exposure and claimable loss, then uses this representation for pricing, control, and risk transfer. It uses deterministic economic labels rather than an LLM judge. In our trace-to-loss testbed, trace-economic pricing reduces pricing MAE from $17.7K to $569 and removes regressive cross-subsidy. A 300-trace expert audit accepts 295 labels unchanged. On 1,000 real SWE-smith traces, trace-conditioned controls reduce CVaR$_{95}$ by 72\%. Theorem~1 gives a finite-sample scope condition. We release code, labels, and audit sheets: \url{https://anonymous.4open.science/r/agent-insurance}.
\end{abstract}

\section{Introduction}

AI agents are moving from recommendation systems to operational systems. They
can edit code, call tools, modify databases, send messages, and execute
financial or administrative workflows. The value of agents comes from
automation, but a single bad action can create real and irreversible loss. In
2025--2026, AI coding agents deleted production databases, wiped home
directories, and destroyed business-critical data through single tool
calls~\citep{wolak2025,pocketos2026,aiid1152}. These incidents raise an
economic question: when is it worth deploying autonomous agents if their losses
have not been priced in advance?

Current practice does not answer this question. Providers often limit
responsibility through consequential-damage disclaimers, and existing cyber or
software insurance is mainly organized around breaches, outages, and
professional-service errors rather than autonomous execution failures.
Organizations therefore keep human review or human sign-off in the workflow so
that someone remains accountable when an agent makes a costly mistake. This is
reasonable for high-risk actions, but it weakens the economic case for
automation when it becomes the default. The missing piece is a way to measure
agent-caused losses from execution evidence, price them before deployment,
transfer risk through insurance, and trigger human review only when expected
loss justifies the cost.

This paper asks a deliberately narrow question: \emph{when does autonomous AI
deployment become profitable, and what evidence is sufficient to quantify,
price, insure, and control its operational risk?} Our answer is to quantify risk
at the level of monitored \emph{customer-task-trace} episodes, then use the
loss representation to price coverage, trigger controls, and transfer risk
through insurance. Each episode links the customer, task, trace, exposed assets,
and claimable loss under a contract. This episode-level view is necessary
because the same model can be harmless in a test environment, costly in a
production SaaS migration, and catastrophic in a regulated financial workflow.
Agent liability is therefore not ordinary product risk. It is a joint property
of the customer, task, assets, and trace.

\begin{figure}[t]
\centering
\includegraphics[width=0.9\columnwidth]{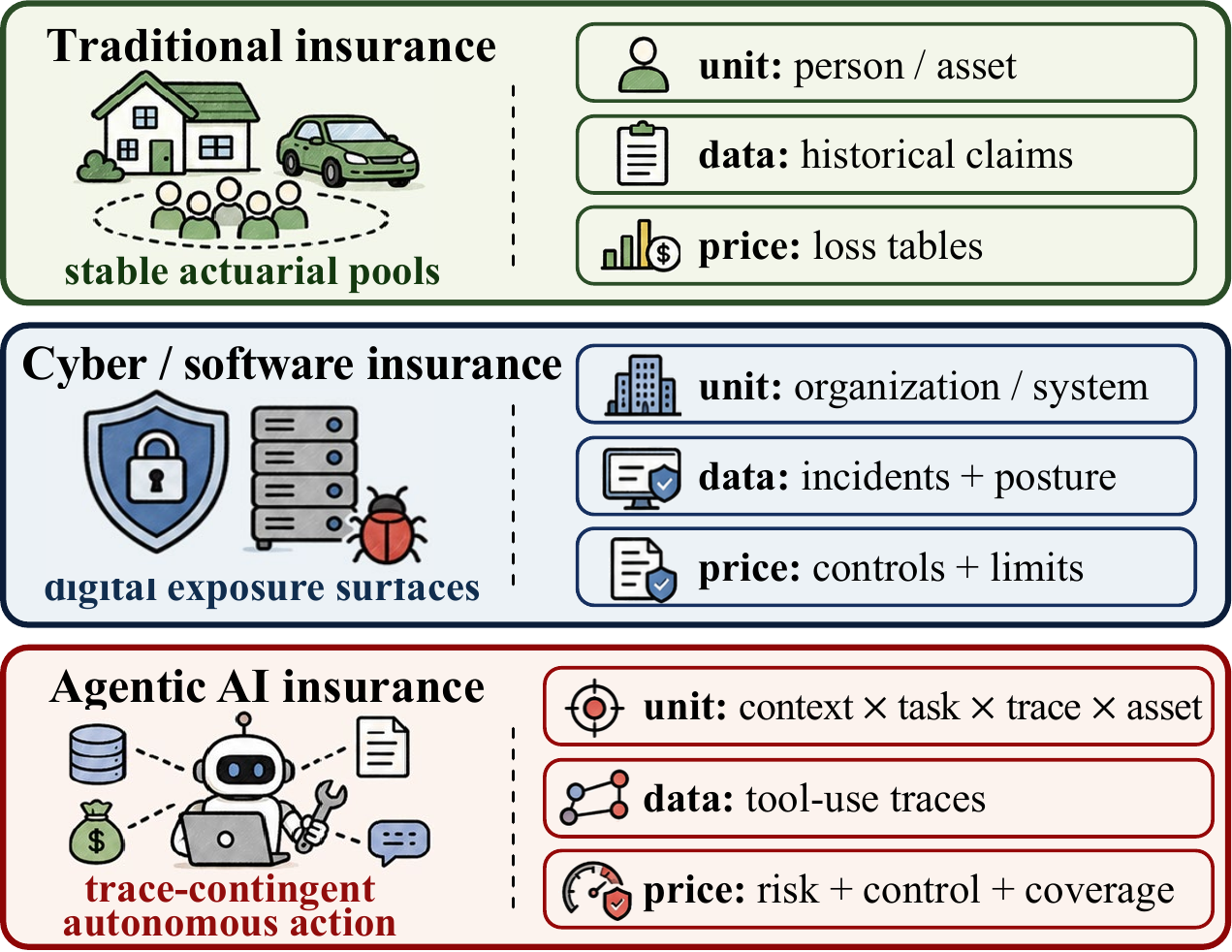}
\vspace{-0.2cm}
\caption{\textbf{Risk-unit shift.} Agentic AI liability cannot be priced from
model identity alone: the same model creates different claimable losses across
customers, tasks, assets, and trace actions, so underwriting must move from
product-level pools to monitored customer-task-trace episodes.}
\vspace{-0.4cm}
\label{fig:insurance_paradigm}
\end{figure}

This risk-unit shift creates three technical challenges. First, the unit of
insurance must be defined. An unrestricted general-purpose agent does not
produce comparable exposures, so product-level pooling hides the heterogeneity
that an insurer must price. Second, labels must be economic rather than purely
semantic. Insurability depends on asset value, claimability, attribution,
deductible, limit, and intervention cost, not only on whether an action name
looks dangerous. Third, pricing must connect to control. A premium transfers
loss, but review, sandboxing, or refusal can prevent loss before an irreversible
action commits. A useful framework must connect trace evidence to both
\emph{loss transfer} and \emph{loss prevention}.

We address these challenges with \emph{trace-economic underwriting}. The core
idea is to make the insurable unit a monitored \emph{customer-task-trace}
episode under a \emph{defined role}. A defined role fixes the task category,
bounds permissions, and produces comparable histories. Each episode maps agent
behavior to economic loss objects, including customer exposure, asset value,
claimable loss, and intervention cost. The method is not an insurance product
or a task-success predictor. It is an underwriting representation that turns
trace evidence into pricing, pre-loss control, and risk-transfer decisions.

We operationalize this representation with deterministic trace-to-loss rules.
The construction parses logs into action classes, annotates actions with
inspectable behavioral dimensions, and combines the trace signal with customer
economics and contract terms to estimate claimable loss. The labels are
documented rules rather than LLM judgments, so assumptions can be audited or
replaced. The same representation supports counterfactual policy evaluation
across pricing and control policies.

The framework supports empirical and formal claims. On a synthetic portfolio,
trace-economic underwriting reduces pricing MAE from \$17.7K to \$569 and
removes a regressive product-flat subsidy in which low-exposure customers
overpay by \$17K--\$20K per episode while financial deployments receive up to
\$55K in implicit subsidy. A 300-trace expert audit accepts 295 of 300 economic
labels unchanged. On 1,000 real SWE-smith traces, trace-conditioned control
reduces CVaR$_{95}$ by 72\%. Theorem~1 explains the scope: trace pricing is
finite-sample identifiable only when a defined role bounds the trace feature
space. General-purpose agents are not insurable objects by themselves.
Insurability attaches to monitored roles with comparable
histories.

Our main contributions are summarized as follows:

\begin{itemize}
\item \emph{Problem formulation.} We formulate insurable AI autonomy as a
trace-conditioned actuarial problem and identify the monitored
customer-task-trace episode under a defined role as the appropriate unit of
liability, pricing, and control.

\item \emph{Method.} We introduce trace-economic underwriting, an auditable
trace-to-loss representation with pricing, pre-loss control, and risk-transfer
operators that distinguish loss transfer from loss prevention.

\item \emph{Empirical validation.} We link real and synthetic traces to customer
exposure, claimable loss, intervention cost, and economic labels.
Validity is established by a 300-trace expert audit with 295 of 300 labels
accepted, 10,037 VCDB incidents, and 500-draw perturbation tests.
\end{itemize}

\section{Related Work}

\subsection{Agent Evaluation and Safety Benchmarks}
Existing agent benchmarks primarily evaluate task success, interaction quality,
or tool-use competence.  SWE-bench, AgentBench, WebLINX, and $\tau$-bench
measure coding, multi-step tool use, web interaction, and customer-service
workflows
\citep{swebench2024,agentbench2023,weblinx2024,yao2024tau}.  These benchmarks
are essential for measuring capability, but their labels are not actuarial
objects: a pass/fail task score does not encode exposed asset value,
claimability, intervention cost, or tail loss.  Safety and governance
benchmarks move closer to our setting by identifying autonomy and tool-use risk
\citep{trl2025,agentsafe2025}, yet they typically stop at qualitative
or ordinal risk categories.  Trace-economic underwriting is complementary: it
keeps the trace as the unit of observation, but adds the economic layer needed
to map actions to claimable losses, pre-loss controls, and risk-transfer
decisions.

\subsection{AI Liability and Risk Transfer}
Legal and policy work argues that agentic AI may require liability rules,
indemnities, or public backstops
\citep{trout2024liability,trout2024insuring,catastrophic2025liability},
and AI procurement already includes indemnity commitments for generative
systems \citep{microsoft2023commitment}.  Insurance-law analyses map affirmative
coverage, silent exposure, exclusions, and mutual-pool proposals for agentic AI
harms \citep{leung2026insurability,afroze2026sail}.  Agent-specific proposals
add decentralized stake mechanisms for autonomous agents
\citep{hu2025insured} and legal-attribution frameworks that trace culpability
to human actors \citep{mukherjee2026operational}.  These works frame risk
transfer as accountability and supply settlement and adjudication layers, but
they treat the loss model as external to the contract.  We study the
measurement layer beneath them, mapping observable agent behavior, deployment
context, and customer exposure to the loss quantities that insurance contracts
require.

\subsection{Actuarial Pricing for Digital and Algorithmic Risk}
Classical insurance theory studies risk pooling, adverse selection, and the
pricing of uncertain losses \citep{arrow1963uncertainty,rothschild1976equilibrium}.
Cyber insurance extends these questions to correlated digital exposure and
sparse incident data \citep{bohme2010modeling,cyberinsurance2023}.  Recent
actuarial work examines algorithmic insurance portfolios and operations risk
\citep{frees2026algorithmic,liu2026insuring}, while agent-specific proposals
study settlement standards, counterfactual runtimes, authority
frontiers, and contract menus for side-effect-bearing actions
\citep{gu2026quantifying,chen2026counterfactual,chen2026authority,yang2025pact}.


\section{Methodology}

The methodology has 3 parts.  We first define the customer-task-trace
episode as the unit of insurable autonomy and state the scope condition under
which such episodes are comparable.  We then construct an auditable
trace-to-loss representation that maps agent behavior to claimable economic
loss without using LLMs as label judges.  Finally, we define pricing, control,
and risk-transfer operators, and show how the same representation induces
contract clauses, solvency constraints, and finite-sample limits.

\subsection{Problem Formulation}

A customer $u$ uses an agentic AI service to automate task category $c$.
During execution the agent emits a trace
$\tau=(a_1,\ldots,a_T)$ of tool calls and messages. An insurance contract
$h=(D,C,\rho)$ specifies deductible $D$, limit $C$, user retention $\rho$, and
evidence rules for claimability.  An underwriting policy must choose a premium
$P$ and, optionally, a pre-loss control action
$b\in\{\mathrm{allow},\mathrm{review},\mathrm{sandbox},\mathrm{stop}\}$.

The insurable unit is not a model product, but a monitored episode
\[
e_i=(u_i,c_i,\tau_i,V_i,A_i,K_i,L_i),
\]
where $V_i$ is task value, $A_i$ is exposed asset value, $K_i$ is pre-loss
control cost, and $L_i$ is expected claimable loss under $h$.  The customer
profile bridges behavior and economics: the same write action can be harmless
for a read-only analyst, costly for a coding SaaS team, and catastrophic for a
financial-operations workflow.  We instantiate read-only, coding SaaS,
financial-operations, and support-operations profiles so that cross-subsidy
under product-flat pricing becomes measurable rather than rhetorical.  A
defined role fixes the task category, bounds permissions, and accumulates
comparable traces; this is the scope condition under which the episode
distribution is learnable.

\subsection{Trace-Economic Episode Construction}

Algorithm~\ref{alg:episode} constructs each episode through three deterministic
layers.  Layer~1 parses an agent log into action classes such as read, write,
execute, message, database, financial, and delete.  Layer~2 assigns five
inspectable dimensions to each action: irreversibility $\alpha_t$, blast radius
$\beta_t$, epistemic uncertainty $\gamma_t$, temporal position $\delta_t$, and
causal attribution $\epsilon_t$.  The action score and trace aggregate are
\begin{align*}
r_t&=\alpha_t\big(w_\beta\sigma(\beta_t)+w_\gamma\gamma_t+
w_\delta\delta_t+w_\epsilon\epsilon_t\big),\\
R(\tau)&=(1-\kappa)\bar r+\kappa\,\cvar_q(r_{1:T}).
\end{align*}
Irreversibility is a multiplicative gate: a read-only action cannot generate
claimable loss regardless of other dimensions.  Layer~3 maps $R(\tau)$,
customer profile, task category, asset value, and contract terms to claim
probability $p_i$, conditional severity $S_i$, and an action-attribution proxy
$v_i$:
\[
L_i=p_i v_i(1-\rho)\,\E[\min(\max(S_i-D,0),C)] .
\]

\begin{figure}[h]
\begin{small}
\begin{algorithm}[H]
\caption{Trace-Economic Episode Construction}
\label{alg:episode}
\begin{algorithmic}[1]
\Require Agent log $\ell$, customer profile $u$, contract $(D,C,\rho)$, task category $c$
\Ensure Episode $e=(u,c,\tau,V,A,K,L)$ with auxiliary $(p,v,\Delta L)$
\State Parse $\ell$ into $\tau=(a_1,\ldots,a_T)$; record tool, arguments, step, and external-state flag
\For{$t=1$ \textbf{to} $T$}
  \State $\mathcal C_t\leftarrow\mathrm{classify}(a_t)$
    \hfill{\small$\triangleright$ read/write/execute/message/db/financial/delete}
  \State $\alpha_t\leftarrow\mathrm{irreversibility}(\mathcal C_t,a_t,u)$
  \State $\beta_t\leftarrow\mathrm{blast\_radius}(a_t,u)$;
         $\gamma_t\leftarrow\mathrm{uncertainty}(\mathcal C_t)$
  \State $\delta_t\leftarrow t/T$;
         $\epsilon_t\leftarrow\mathrm{attribution}(\mathcal C_t,a_t)$
  \State $r_t\leftarrow\alpha_t(w_\beta\sigma(\beta_t)+w_\gamma\gamma_t
         +w_\delta\delta_t+w_\epsilon\epsilon_t)$
\EndFor
\State $R(\tau)\leftarrow(1-\kappa)\bar r+\kappa\,\widehat{\mathrm{CVaR}}_q(r_{1:T})$
\State $p\leftarrow\sigma(aR(\tau)+b)$
  \hfill{\small$\triangleright$ domain-calibrated two-parameter link}
\State $A\leftarrow u.\mathrm{asset\_value}$;\quad
       $K\leftarrow u.\mathrm{review\_cost}$;\quad
       $V\leftarrow\mathrm{task\_value}(u,c)$
\State $S\leftarrow\mathrm{severity}(R(\tau),c,u,A)$;
       $v\leftarrow\mathrm{mean}(\epsilon_{1:T})$
\State $L\leftarrow p\,v(1-\rho)\,\mathbb E[\min(\max(S-D,0),C)]$
\State $\Delta L\leftarrow L\cdot u.\mathrm{control\_effect}$
\State \Return $e=(u,c,\tau,V,A,K,L)$
\end{algorithmic}
\end{algorithm}
\vspace{-0.8cm}
\end{small}
\end{figure}

All labels are produced by documented rules rather than LLM judges.  This makes
severity tables, profile assumptions, deductibles, limits, and attribution
discounts auditable and replaceable.

\begin{figure*}[t]
\centering
\includegraphics[width=0.98\textwidth]{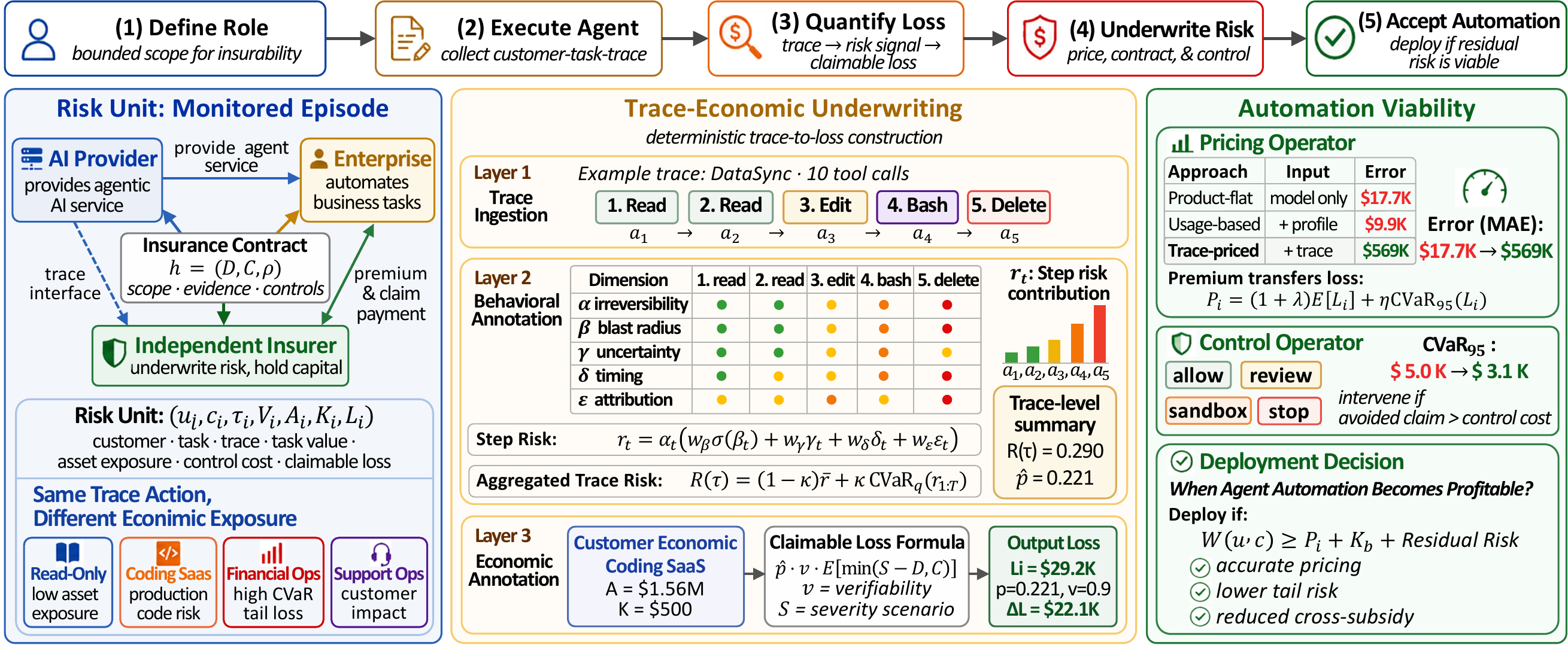}
\vspace{-0.15cm}
\caption{\textbf{Framework overview.}
\textit{Left}: a customer uses an agentic AI service for task automation, while an
independent insurer underwrites the
trace-based contract. \textit{Center}: the trace-economic construction parses
logs, annotates behavioral dimensions, and combines the resulting trace signal
with customer economics to produce claimable loss. \textit{Right}:
product-flat pricing pools heterogeneous customers, whereas trace-conditioned
pricing and control use the customer-task-trace episode as the risk unit.}
\vspace{-0.4cm}
\label{fig:method_pipeline}
\end{figure*}

The construction separates live trace risk from economic calibration.  In the
experiments, completed historical traces support rating, while only the current
trace prefix can trigger control under a precommitted contract.  The five
dimensions are label-free and computable before each irreversible action; the
two-parameter link from $R(\tau)$ to failure probability is
domain-specific and can be recalibrated from a small labeled set.  This matters
because action engagement is not universally harmful: in coding tasks, a high
action tail often indicates operational risk, while in customer-service tasks a
low-action trace may fail because policy constraints prevent the agent from
acting.  The representation therefore evaluates the whole trace-economic
episode, not a universal scalar risk score.

\paragraph{Auditability and portability.}
The construction is deliberately modular.  Layer~1 can be replaced by a parser
for web, database, workflow, or customer-service agents; Layer~2 keeps the same
risk dimensions; Layer~3 can replace scenario severities with insurer claim
data when such data become available.  The policy interface does not change:
every parser must output the same episode object and auxiliary quantities
$(p,v,\Delta L)$.  This is why the empirical instantiation is not a
synthetic-label exercise alone.  It specifies which assumptions are behavioral,
economic, and contractual, so that a later deployment can audit or recalibrate
one layer without rewriting the underwriting problem.

\paragraph{Risk-signal validation.}
Before converting traces into dollars, we verify that the trace dimensions rank
operational danger.  On 5,000 SWE-smith trajectories, the interpretable
five-dimensional score with only a two-parameter calibration reaches AUC 0.637.
A GRU over raw tool-call sequences reaches AUC 0.676, and a bag-of-tools
logistic model reaches 0.658.  Both trained baselines can run online, but the
sequence model is not decomposable into claim-relevant causes and is difficult
to justify to customers or claims adjusters.  Our signal recovers 78\% of the
GRU's AUC improvement above random while remaining incremental and auditable.
A leave-one-out ablation further shows
that irreversibility is the dominant dimension: removing $\alpha$ drops
Spearman rank correlation from 0.948 to 0.751, whereas removing blast radius or
epistemic uncertainty causes smaller independent losses.  Thus the framework
does not claim that the hand-built score is the best predictor; it claims that
the score is a defensible underwriting signal whose assumptions can be inspected
and stress-tested.

The calibration requirement is intentionally weak.  The risk dimensions are
label-free, but the map from trace risk to failure probability is
domain-specific.  We therefore fit only $\hat p=\sigma(aR(\tau)+b)$ rather than
a learned sequence model. This keeps the direction and scale of risk
recalibratable while preserving the same feature definitions, and
prevents the method from optimizing for task failure when the insurance
object is claimable loss under a contract.

\subsection{Pricing, Control, and Risk Transfer}

We evaluate policies by their information set.  Product-flat pricing uses only
model/product identity; usage-based pricing adds customer profile and task
category; trace-only pricing uses $R(\tau)$ without customer economics; and
trace-priced underwriting uses the full customer-task-trace episode.  Static
tool control intervenes whenever a dangerous tool class appears; trace control
intervenes only when expected avoided claim exceeds review or sandbox cost.
Premiums combine expected loss and tail loading:
\[
P_i=(1+\theta)\widehat{\E[L_i]}+\lambda\,\widehat{\cvar}_{95}(L_i).
\]
For customer $u$, task $c$, trace $\tau$, and contract $h$, the common object
behind pricing, control, and contract design is the trace-conditional loss
surface
\begin{align*}
\mathcal L_\theta(u,c,\tau;h)
&=\big(p_\theta,F_\theta,a_\theta,K_\theta\big),\\
Y_h&=a_\theta(1-\rho)\min\{(S-D)_+,C\}.
\end{align*}
Here $p_\theta$ is claim probability, $F_\theta$ is conditional severity,
$a_\theta$ is attribution/verifiability, and $K_\theta$ is pre-loss review or
sandbox cost.  The economically optimal control under information $\mathcal G$
is
\[
\Gamma(\mathcal L\mid\mathcal G)
\in\arg\min_{b\in\mathcal B(\tau)}\{\E[Y_h^b\mid\mathcal G]+K^b\}.
\]
For a binary allow/intervene decision, intervention is optimal exactly when
\[
\E[Y_h^{\mathrm{allow}}-Y_h^{\mathrm{int}}\mid\mathcal G]
\ge K^{\mathrm{int}}-K^{\mathrm{allow}} .
\]
This separates pricing, which transfers losses, from control, which prevents
them, and explains why a static dangerous-tool blacklist is optimal only when
tool class is already sufficient for avoided claim net of friction.

\subsection{Scope, Contracts, and Insurability}

\begin{figure*}[t]
\centering
\includegraphics[width=0.98\textwidth]{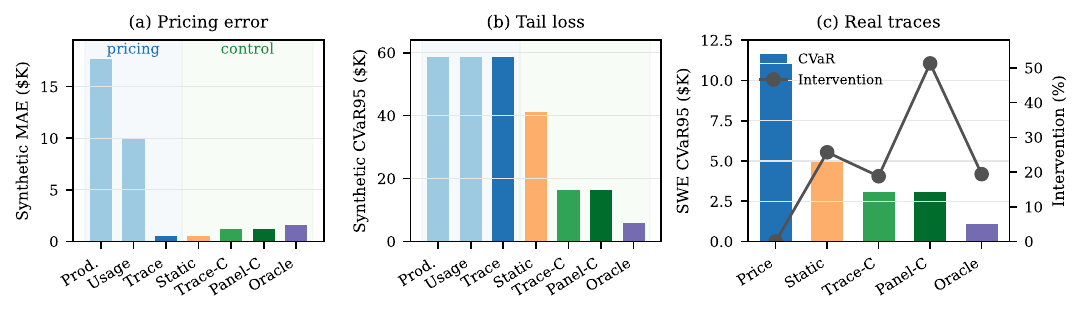}
\vspace{-0.55cm}
\caption{\textbf{Pricing versus control.}  Trace information makes premiums
accurate, but accurate premiums only transfer losses; tail risk falls only when
the same trace signal triggers pre-loss control, matching AI-panel protection
with less review on real SWE-smith traces.}
\vspace{-0.35cm}
\label{fig:pricing_control_progression}
\end{figure*}

The loss surface also supports contract clauses.  Deductibles, limits,
coinsurance, exclusions, sublimits, and reinsurance are not post-hoc legal
details: they change $Y_h$, $a_\theta$, tail capital, and the feasible control
set.  A high-verifiability database write can receive ordinary indemnity;
partial-verifiability model-output claims may require coinsurance and
sublimits; common model-update failures require an explicit systemic layer.  If
\[
Y_i=(1-Z_0)Y_i^{\mathrm{id}}+Z_0Y_i^{\mathrm{sys}},
\qquad Z_0\sim\mathrm{Bernoulli}(q),
\]
then diversification does not eliminate portfolio risk because
\[
\Var\!\left(\frac1N\sum_{i=1}^N Y_i\right)
\to q(1-q)(\E[Y^{\mathrm{sys}}]-\E[Y^{\mathrm{id}}])^2 .
\]
Thus trace segmentation must be paired with limits, reinsurance, exclusions, or
a public backstop for systemic loss.

These operators yield an episode-level insurability gap.  For buyer value
$W(u,c)$ and action $b$, define
\begin{align*}
G_b(\mathcal G)
&=\E[Y_h^b\mid\mathcal G]
  +\lambda\,\cvar_\alpha(Y_h^b\mid\mathcal G)\\
&\quad +K^b+c_{\mathrm{admin}}-W(u,c),
\end{align*}
and $G^\star(\mathcal G)=\min_{b\in\mathcal B(\tau)}G_b(\mathcal G)$.
A deployment is privately viable only when $G^\star\le0$.  Control therefore
does more than lower a risk score: it can move an episode across the viability
boundary by reducing expected claim and tail capital more than it adds review
cost.

\paragraph{Proposition 1 (value of trace information).}
Let $X$ be product identity and $Z=(X,U,\tau)$ add customer and trace
information.  For claimable loss $Y$, moving from product pricing to
trace-conditional pricing reduces optimal squared pricing error by
\begin{align*}
&\E[(Y-\E[Y\mid X])^2]-\E[(Y-\E[Y\mid Z])^2]\\
&\qquad=\E[\Var(\E[Y\mid Z]\mid X)].
\end{align*}
The gain is positive exactly when customers or traces with the same product have
different conditional claim costs.

\paragraph{Theorem 1 (finite-sample scope condition).}
Let $(X_i,Z_i,Y_i)_{i=1}^n$ be iid episodes with $Y_i\in[0,B]$ and feature map
$\phi(Z)\in\mathbb R^d$, $\|\phi(Z)\|_2\le R$.  Assume the product and
trace-conditional means lie in nested bounded linear classes.  For their
empirical risk minimizers, let $\mathcal E_X$ and $\mathcal E_Z$ denote
population squared risks.  Then, with probability at least $1-\delta$,
\[
\mathcal E_X-\mathcal E_Z
\ge \Delta_{\mathrm{info}}
-C B^2R^2\sqrt{\frac{d\log n+\log(1/\delta)}{n}},
\]
where $\Delta_{\mathrm{info}}=\E[\Var(\E[Y\mid Z]\mid X)]$.  Hence trace pricing
is identifiable once
\begin{equation}
n^\star=\tilde\Omega\!\left(\frac{B^4R^4d}{\Delta_{\mathrm{info}}^2}\right).
\label{eq:sample_complexity}
\end{equation}
A defined role bounds $d$ through bounded permissions and trace length.  For a
role $V=(\mathcal M(v),\mathcal T_c,\mathcal F,H)$, one obtains
$d\le|\mathcal F|T_{\max}$ and a finite identification threshold.  An
unrestricted general-purpose agent lacks this verifiable scope: any finite
feature compression is a modeling choice an insurer cannot audit or defend as a
contract boundary.  The insurability claim is therefore not that prediction is
impossible in principle, but that trace pricing becomes actuarially defensible
only after the role bounds the feature space and produces comparable histories.
\emph{Proof sketch.}
Proposition~1 is the law of total variance applied to the Bayes pricing rule
$m_{\mathcal G}=\E[Y\mid\mathcal G]$.  Theorem~1 subtracts finite-sample
estimation error from this Bayes-risk gap, so the lower bound becomes positive
only when comparable histories are large enough relative to the role-bounded
trace dimension.

\begin{table}[t]
\centering
\caption{\textbf{Evidence stack.}  Because closed AI-agent claim histories do
not yet exist, the empirical validation triangulates insurability evidence from synthetic
portfolios, real tool-use traces, loss taxonomies, monetary anchors, expert
review, and stress tests.}
\vspace{-0.3cm}
\label{tab:evidence_stack}
\small
\setlength{\tabcolsep}{2.4pt}
\begin{tabular}{llrl}
\toprule
Component & Source & Scale & Role \\
\midrule
Portfolio & simulator & 25K eps. & controlled tests \\
Real traces & SWE-smith & 1K traj. & tool-use behavior \\
Loss channels & VCDB & 10,037 inc. & coverage audit \\
Severity anchors & public cases & 31 & dollar scale \\
Expert audit & manual review & 300 traces & label validity \\
Stress audit & perturbations & 500 draws & sensitivity \\
\bottomrule
\end{tabular}
\vspace{-0.4cm}
\end{table}

\begin{figure*}[t]
\centering
\includegraphics[width=0.98\textwidth]{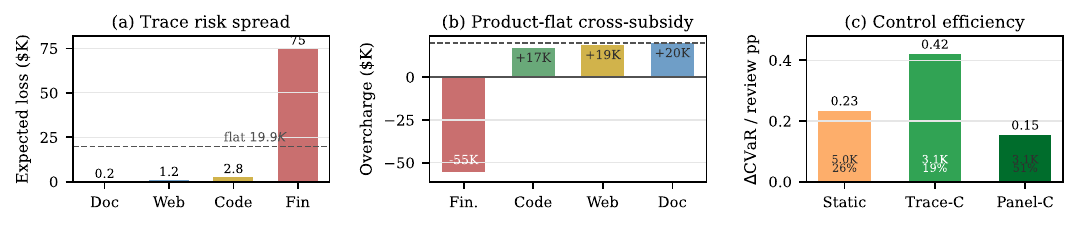}
\vspace{-0.45cm}
\caption{\textbf{Trace mechanism.}  The customer-task-trace episode is the
insurable unit: it reveals heterogeneous exposure, shows how product-flat
pricing converts heterogeneity into cross-subsidy, and identifies which trace
conditions make control efficient.}
\vspace{-0.35cm}
\label{fig:trace_mechanism}
\end{figure*}

\emph{Feature compression and auditability.}
Embeddings or regularization can compress a general-purpose trace space into a
finite representation, but compression alone does not define an insurable
contract.  A role specification tells an insurer which tools, permissions, task
states, and failure modes belong in the feature map.  Without that scope, a
claims adjuster cannot verify why a trace was placed in one risk cell rather
than another, and a customer cannot know which activities the premium
covers.  The scope condition is therefore a contract-design requirement as much
as a statistical one.

\section{Experiments}
\label{sec:results}

\paragraph{Protocol.}
The experiments evaluate whether trace-economic underwriting satisfies the
requirements for insurable autonomy: accurate pricing, pre-loss control,
transferable risk signals, private viability, insurer solvency, and stable
pooling.  We use a synthetic portfolio with 5 seeds $\times$ 5,000 episodes
across 4 customer profiles and four task categories (financial, coding,
document, and web/support), and 1,000 real
SWE-smith coding-agent trajectories mapped to the same episode schema.  Product
pricing uses only model identity; usage pricing adds customer and task groups;
trace pricing conditions on the monitored episode; and control policies
intervene when avoided claim and tail capital exceed review friction.  We
report pricing MAE, residual CVaR$_{95}$, intervention burden, audit evidence,
and the insurability gap in this section.  Table~\ref{tab:evidence_stack} summarizes the evidence stack; the
remaining displays test pricing, control, transfer, insurability, solvency, and
market effects.  Appendix~\ref{app:schema} gives implementation
details, including customer profiles, parser outputs, action rubrics, and
economic labels for reproduction.

\subsection{Pricing and Control}

Figure~\ref{fig:pricing_control_progression} asks whether trace information
solves both pricing and risk prevention.  In the synthetic portfolio,
product-flat pricing pools heterogeneous exposures and has MAE \$17.7K; usage
pricing helps but still omits the realized trace; trace pricing reduces MAE to
\$569.  Yet all pricing-only policies leave residual CVaR unchanged because
premiums transfer losses rather than prevent them.  Tail reduction appears only
after control: on SWE-smith, trace control matches AI-panel control at \$3.1K
CVaR$_{95}$ while reviewing 18.8\% rather than 51.3\% of episodes.  The result
separates two design goals that are often conflated: actuarial pricing and
pre-loss harm prevention.  Appendix~\ref{app:proofs} gives the full algebra
behind these pricing and control operators.

\begin{table}[t]
\centering
\caption{\textbf{Out-of-profile transfer.}  Trace-economic pricing is not just
memorizing customer groups: when calibrated without one customer profile and
tested on that held-out profile, it still sharply reduces pricing MAE on real
SWE-smith traces, including the hardest financial profile.}
\vspace{-0.3cm}
\label{tab:profile_transfer}
\small
\setlength{\tabcolsep}{2.2pt}
\begin{tabular}{lrrrr}
\toprule
Held-out & $n$ & Product & Trace & Red. \\
\midrule
Coding & 709 & \$4,067 & \$271 & 93.3\% \\
Financial & 41 & \$6,080 & \$1,890 & 68.9\% \\
Read-only & 162 & \$2,558 & \$10 & 99.6\% \\
Support & 88 & \$2,322 & \$154 & 93.4\% \\
\bottomrule
\end{tabular}
\vspace{-0.4cm}
\end{table}

Figure~\ref{fig:trace_mechanism} isolates why trace conditioning is needed.
Panel (a) shows that expected loss varies sharply across customer-task traces;
Panel (b) shows that product-flat pricing converts this heterogeneity into
cross-subsidy; Panel (c) shows that trace control buys more tail reduction per
review point than static or panel-heavy control.  The result explains why the
risk unit cannot be model identity or trace alone: insurable exposure is
created by the interaction between customer context, task, and realized action
trace.

\subsection{Evaluator Validity and Robustness}

The validity question is whether a scenario-calibrated evaluator can support
policy comparison before closed-claims data exist.  A 300-trace expert audit
accepts 295 of 300 labels unchanged, confirming the labeling rules are
defensible.  Table~\ref{tab:evidence_stack} shows the full evidence stack: real
SWE-smith traces supply behavior, VCDB audits 10,037 incidents for channel
coverage, and public cases anchor dollar scale.  Five hundred perturbation
draws re-evaluate the comparison under bounded label and parser changes.  This
is a sensitivity analysis, not a formal robustness certificate or a claim of
final actuarial rates.  The release contains the expert and panel sheets;
Appendix~\ref{app:stress} reports stress and transfer tests, and
Appendix~\ref{app:dim_ablation} reports risk-dimension ablations.

Table~\ref{tab:profile_transfer} adds a transfer test on real traces.  For
each customer profile, the trace-economic price is calibrated on the other
profiles and evaluated on the held-out one.  Product-flat pricing has no way to
adapt to the omitted profile; trace pricing still uses task value, asset
exposure, and realized action risk.  The held-out financial profile is the
hardest case because it has the smallest sample and the heaviest severity tail,
yet trace pricing still reduces MAE by 68.9\%.  The result supports the
framework's main modeling claim: reusable information lies in the joint
customer-task-trace structure, not in memorized customer labels.  Appendix~\ref{app:stress}
gives the full transfer table with usage-based baselines and the perturbation
audit.

\subsection{Insurability and Market Effects}

\begin{table}[t]
\centering
\caption{\textbf{Insurability sensitivity.}  Control affects the private
coverage boundary, not only safety: as tail capital becomes expensive, the
best trace-aware control can move episodes from uninsurable to viable, although
control friction can reduce viability when tail loading is zero.}
\vspace{-0.3cm}
\label{tab:tail_sensitivity}
\small
\setlength{\tabcolsep}{2.2pt}
\begin{tabular}{llrrr}
\toprule
Dataset & $\lambda$ & Allow & Best & Gap closed \\
\midrule
Synthetic & 0.0 & 100.0\% & 91.6\% & \$2,191 \\
Synthetic & 0.3 & 78.7\% & 79.1\% & \$2,488 \\
Synthetic & 1.0 & 51.5\% & 67.5\% & \$3,180 \\
SWE-smith & 0.0 & 100.0\% & 98.4\% & \$259 \\
SWE-smith & 0.3 & 95.9\% & 96.6\% & \$297 \\
SWE-smith & 1.0 & 83.6\% & 83.9\% & \$385 \\
\bottomrule
\end{tabular}
\vspace{-0.4cm}
\end{table}

Table~\ref{tab:tail_sensitivity} asks whether control changes private
insurability, not only safety.  We evaluate $G^\star$, which combines expected
claim, tail capital, control cost, and buyer value.  At the base loading, the
effect is modest; when tail capital is expensive ($\lambda=1.0$), synthetic
viability rises from 51.5\% to 67.5\%.  Control is therefore a market-design
operator: it can move deployments across the private-coverage boundary by
reducing tail capital more than it adds review cost.  Appendix~\ref{app:viability}
gives the full viability frontier and sensitivity plots.

\begin{figure}[t]
\centering
\includegraphics[width=0.90\columnwidth]{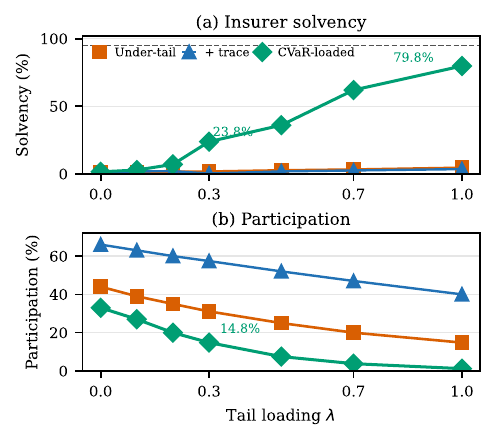}
\vspace{-0.45cm}
\caption{\textbf{Portfolio tradeoff.}  Trace segmentation improves risk
measurement, but solvency requires explicit tail capital; CVaR-loaded premiums
protect the insurer at the cost of lower buyer participation.}
\vspace{-0.35cm}
\label{fig:portfolio_tradeoff}
\end{figure}

Figure~\ref{fig:portfolio_tradeoff} tests a negative hypothesis: perhaps better
trace segmentation alone makes the insurer solvent.  It does not.  Under-tail
pricing remains insolvent even with trace information, while CVaR-loaded pricing
raises solvency from 1.7\% to 23.8\% at $\lambda=0.3$ and from 4.5\% to 79.8\%
at $\lambda=1.0$.  Here solvency is one-period premium sufficiency: collected
premium covers realized claims and administrative cost, without outside capital
or reinsurance.  The cost is lower participation.  The result shows that a
better $\phat$ is not enough; the premium formula must explicitly carry tail
capital, and systemic layers still require limits, reinsurance, or backstops.
Appendix~\ref{app:portfolio} reports portfolio diagnostics and
Appendix~\ref{app:contracts} expands the contract clauses implied by this
tradeoff.

Figure~\ref{fig:adverse_selection} studies the market consequence of the
cross-subsidy in Figure~\ref{fig:trace_mechanism}.  Low-risk customers are
overcharged under product-flat pricing, so they exit; the residual pool becomes
increasingly financial, and the flat premium rises toward the high-risk
actuarial cost.  This turns the fairness problem into a market-stability
problem: product-flat pricing is not merely inaccurate, it can destroy the
low-risk side of the market.  Trace-conditioned underwriting removes this
spiral by charging each monitored episode according to its own exposure.

\paragraph{Threats to validity.}
The strongest threat is that severity labels are scenario-calibrated from public
incidents, not closed insurer claims.  A 300-trace expert audit (295/300
accepted) confirms label validity, but not magnitude.  The results
should therefore be read as policy-ordering and market-design evidence, not
final actuarial rates.  A second threat is domain imbalance: SWE-smith supplies
real tool-use traces but is mostly coding, while high-stakes insurance demand
may be strongest in finance, healthcare, and support operations.  A third threat
is intervention-cost calibration; review friction varies by organization and
task urgency.  The released implementation exposes these assumptions so future
claim histories can replace them rather than change the underwriting interface.

\section{Discussion}

\begin{figure}[t]
\centering
\includegraphics[width=0.90\columnwidth]{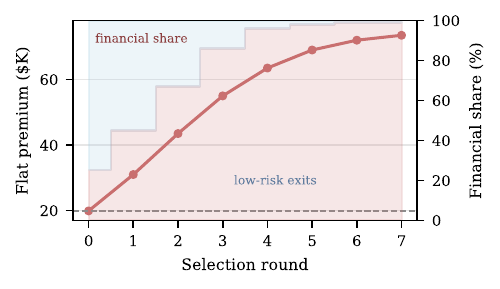}
\vspace{-0.45cm}
\caption{\textbf{Adverse selection.}  Product-flat pricing creates a market
failure: low-risk customers exit after subsidizing high-risk deployments, the
pool becomes increasingly financial, and the flat premium rises toward
high-risk cost.}
\vspace{-0.35cm}
\label{fig:adverse_selection}
\end{figure}

\paragraph{Social impact.}
Figures~\ref{fig:trace_mechanism} and~\ref{fig:adverse_selection} show that
product-flat AI-agent insurance is not only inaccurate but regressive:
lower-risk document-processing and web-automation customers subsidize
financial-operations deployments by \$17K--\$20K per episode, then exit the
pool.  Trace-economic underwriting links coverage to auditable pre-loss
controls and removes this cross-subsidy.  This matters because human oversight
should be allocated where it reduces expected claim and tail loss, not retained
as a default liability shield.  Without credible risk transfer, agent-caused
losses remain with users and residual human overseers rather than funded
accountability mechanisms.

\paragraph{Auditability.}
Insurance claims require stable, contestable rules, so the implementation uses
deterministic economic labels while reserving LLMs or humans for review and
auditing.  The target itself should be inspectable by customers, insurers,
claims adjusters, and regulators.  Appendix~\ref{app:calibration} gives the
severity anchors, calibration provenance, and incident checks that make this
rule set auditable and replaceable.

\paragraph{Limitations.}
The framework applies to role-bounded work with specified tasks, permissions, and loss channels, such as translation, compliance review, coding, or audits. It is not a pricing method for open-ended exploration, self-improving loops, or agents that create new goals during deployment. In those settings, the action space and loss surface lack comparable histories, so coverage should wait for narrower roles, explicit caps, or public-risk governance rather than ordinary underwriting.

\section{Conclusion}

We introduced trace-economic underwriting to characterize when autonomous AI
deployment becomes profitable through quantified and insured risk. AI liability
should not be priced at the product level but at the customer-task-trace level,
as economic loss depends jointly on what the agent did and where it did it.
This framework is tractable only when agents operate in defined roles with fixed
task categories, bounded permissions, and comparable execution traces. Without
such a role, trace pricing is not finite-sample identifiable, and
general-purpose agents fall outside this regime. In both synthetic portfolios
and real SWE-smith traces, trace-economic pricing removes cross-subsidy, while
trace-aware control reduces tail risk. The result is a path toward agent
deployment where failure need not be eliminated before use, but must be
quantified, insured, and tied to bounded liability rather than broad
indemnity promises.

\bibliography{refs}

\clearpage
\appendix
\section{Appendix: Reproducibility Details}

\subsection{Repository Structure}

The trace-economic underwriting implementation is under
\texttt{experiments/trace\_econ/}.  The main scripts are:

\begin{itemize}
\item \texttt{build\_dataset.py}: builds synthetic, SWE-smith, and tau-bench
episodes in a common trace-economic schema.
\item \texttt{risk\_model.py}: deterministic trace-to-economic-risk labeler.
\item \texttt{run\_experiments.py}: evaluates pricing and control policies.
\item \texttt{run\_multiseed.py}: runs the synthetic multi-seed experiment.
\item \texttt{run\_sensitivity.py}: perturbs severity, review cost, and tail
loading.
\item \texttt{analyze\_review\_cost\_stress.py}: doubles financial-profile
and all-profile review costs on real SWE-smith traces.
\item \texttt{make\_human\_audit\_sheet.py}: creates the external audit sheet.
\item \texttt{expert\_preannotate\_audit.py}: creates the expert
pre-annotation and second-review sheet.
\item \texttt{finalize\_expert\_review.py}: records the second-expert reviewed
audit after reviewer confirmation.
\item \texttt{external\_anchors/}: maps VCDB incidents into trace-economic
loss channels and summarizes monetary severity anchors.
\end{itemize}

\subsection{Episode Schema}
\label{app:schema}

Each episode stores:
\begin{align*}
e=\{&\texttt{episode\_id}, u, c, V, A, T,\texttt{resolved},\\
&R(\tau),p,S,\E[L],\E[\mathrm{Claim}],K,\Delta L_K\}.
\end{align*}
Here $u$ is customer profile, $c$ is task category, $V$ is task value, $A$ is
asset value, $T$ is trace length, $R(\tau)$ is trace risk, $p$ is loss
probability, $S$ is conditional severity, $K$ is intervention cost, and
$\Delta L_K$ is expected claim reduction if a pre-loss control is applied.

Each step stores:
\begin{align*}
s_t=\{&\texttt{episode\_id},t,\texttt{tool},\texttt{action\_type},\\
&\alpha,\beta,\gamma,\delta,\epsilon,r_t,\\
&\texttt{step\_exposure},\texttt{preventable}\}.
\end{align*}

\subsection{Structural Challenges and Prior-Work Coverage}

Four structural challenges organize the positioning of the underwriting
framework.  We summarize them compactly here.
Tables~\ref{tab:structural_challenges_app} and~\ref{tab:related_matrix_app}
show why the underwriting framework must combine trace observation, pricing, dynamic
control, systemic exposure, and attribution rather than treating AI-agent
insurance as ordinary product insurance.

\begin{table}[h]
\centering
\caption{\textbf{Structural gap.}  Agentic AI insurance differs from ordinary
product insurance because risk is endogenous to deployment context, changes
during execution, can arrive as common shocks, and requires trace-verifiable
causation; the underwriting response is to price and control monitored episodes.}
\label{tab:structural_challenges_app}
\small
\setlength{\tabcolsep}{2pt}
\begin{tabular}{p{0.17\linewidth}p{0.36\linewidth}p{0.37\linewidth}}
\toprule
Challenge & Why prior insurance assumptions fail & Underwriting response \\
\midrule
Endog. $p$ & No stable historical class exists for model version, task,
and deployment context & Estimate risk from observed traces, then recalibrate
severity by domain \\
Common shock & A model update can shift all policyholders' failure rates at
once & Explicit systemic layer and reinsurance/backstop condition \\
Dynamic exec. & Risk is observable before an irreversible action occurs &
Step-level scoring and economically justified controls \\
Cause & Prompt injection, model changes, and context drift make
claims contestable & Trace verifiability, exclusions, warranties, audit fields \\
\bottomrule
\end{tabular}
\end{table}

\begin{table}[h]
\centering
\caption{\textbf{Related-work positioning.}  Existing work covers parts of the
problem---safety traces, liability theory, cyber pricing, or systemic risk---but
does not jointly provide trace-conditioned pricing, dynamic control, systemic
stress, and attribution fields for agent insurance.}
\label{tab:related_matrix_app}
\small
\setlength{\tabcolsep}{2pt}
\begin{tabular}{p{0.24\linewidth}ccccc}
\toprule
Work cluster & Trace & Price & Sys. & Dyn. & Attr. \\
\midrule
AI liability policy & No & No & Partial & No & Partial \\
PRA / scenario risk & No & No & Partial & No & No \\
Agent safety benchmarks & Yes & No & No & Partial & No \\
ARS settlement & Partial & Partial & No & No & No \\
Static algorithmic liability & No & Yes & Partial & No & Partial \\
Cyber insurance & No & Yes & Partial & No & Partial \\
\textbf{This work} & \textbf{Yes} & \textbf{Yes} & \textbf{Yes} &
\textbf{Yes} & \textbf{Yes} \\
\bottomrule
\end{tabular}
\end{table}

\subsection{Customer Profiles}

\begin{table}[h]
\centering
\caption{\textbf{Customer economics.}  The same trace action has different
insurance meaning across customers; these profiles attach asset exposure,
production use, and review cost before claimable loss is computed.}
\label{tab:customer_profiles_app}
\small
\setlength{\tabcolsep}{3pt}
\begin{tabular}{lccc}
\toprule
Profile & Mean asset & Production & Review cost \\
\midrule
read-only & \$2K & 0.05 & \$20 \\
coding SaaS & \$35K & 0.35 & \$80 \\
financial ops & \$250K & 0.65 & \$220 \\
support ops & \$12K & 0.45 & \$45 \\
\bottomrule
\end{tabular}
\end{table}

Table~\ref{tab:customer_profiles_app} is the economic bridge between traces and
losses.  Profiles also differ in rollback quality and task mix.  For example,
read-only customers mostly delegate document and support lookups, while
financial-operations customers expose higher asset values and have higher
review costs.  Profiles also assume a single agent executes the task;
multi-agent decompositions \citep{xu2026multi} would require a composite
profile or per-agent endorsements.

\subsection{Action Classification}

Tool calls and arguments are deterministically mapped to action classes:
read, validation, write, execute, message, support-change, database, financial,
and delete.  Regular expressions identify common shell commands, database
operations, financial terms, support operations, messaging actions, and file
paths.  This design makes parser errors visible: the audit explicitly marks
cases where a coding trace may be classified as financial because of lexical
overlap.

\subsection{Annotation Rubric Summary}

Table~\ref{tab:annotation_rubric_summary} is the compact annotation rubric used
by the implementation.  It is meant to be read as a reproducibility object:
every label has a target, a scale, a source of evidence, a deterministic rule,
and a dataset field.  Later sections expand the individual rows.

\begin{table*}[t]
\centering
\caption{\textbf{Annotation rubric.}  The implementation separates observable
trace labels from economic labels and records the evidence, rule, and dataset
field for each one, making the trace-to-loss mapping auditable rather than
judge-only.}
\label{tab:annotation_rubric_summary}
\small
\setlength{\tabcolsep}{2.5pt}
\begin{tabular}{p{0.15\linewidth}p{0.16\linewidth}p{0.17\linewidth}p{0.29\linewidth}p{0.13\linewidth}}
\toprule
Target & Values / scale & Evidence used & Rule & Dataset field \\
\midrule
Action class & read, validation, write, execute, message, support-change,
database, financial, delete & tool name and arguments; domain keywords &
deterministic parser maps each tool call to one action class & action type \\
Irreversibility & $\alpha_t\in[0,1]$ & tool side-effect, rollback path,
transaction reversibility & read/dry-run gets 0; external state changes,
messages, financial transactions, and deletes increase $\alpha_t$ &
\texttt{alpha} \\
Blast radius & $\beta_t\ge0$ & files, wildcard scope, database/financial
scope hints, production exposure & log-scaled affected-artifact count plus
domain scope hint & blast radius; exposure \\
Uncertainty & $\gamma_t\in[0,1]$ & self-consistency when available; action
class prior otherwise & high for execute/database/financial/delete; single-run
public traces mark this as weakly measured & \texttt{uncertainty} \\
Timing & $\delta_t\in[0,1]$ & step index and remaining horizon & earlier
irreversible actions receive larger compounding-risk weight & \texttt{temporal} \\
Attribution & $\epsilon_t\in[0,1]$ & log evidence, counterfactual causation,
claim verifiability & direct writes/transactions/messages receive high scores;
advice-only or multi-cause harms are discounted & \texttt{attribution} \\
Trace risk & $R(\tau)\in[0,1]$ & action risk sequence & blend mean action risk
with tail CVaR so one dangerous action can dominate a session &
\texttt{trace\_risk} \\
Customer exposure & dollars, profile class & profile task mix, production
ratio, rollback quality, asset-value prior & attach asset value and review
cost to the trace before computing loss & customer profile; asset value \\
Claimable loss & dollars & severity anchors, deductible, limit, attribution &
expected loss is reduced by deductible, capped by limit, and discounted by
attribution/verifiability & claimable loss \\
Control value & dollars and binary rule & review cost and preventability &
intervene when expected avoided claim exceeds review/sandbox cost &
review cost; avoided loss \\
\bottomrule
\end{tabular}
\end{table*}

\subsection{Synthetic Episode Generative Model}
\label{app:generative_model}

We describe the joint generative process explicitly, addressing the question
of how $R(\tau)$ relates to $p_{\text{true}}$ and whether $\hat{p}$ is a
biased or unbiased estimator.

\textbf{Step 1: latent failure probability.}
For each episode, a task profile $k\in\{\text{financial, coding, document, web}\}$
is drawn with equal probability.  The latent (unobserved) failure probability is
\[
p_{\text{true}}\;\sim\;\mathrm{Beta}(\alpha_k,\beta_k),
\]
with parameters calibrated to published benchmark failure rates
(Table~\ref{tab:scenario_params_app}).  This is the actuarial ground truth that
\emph{no} pricing policy can directly observe.

\textbf{Step 2: synthetic trace generation.}
Action-level risk features are drawn conditionally on $p_{\text{true}}$:
\[
\mathrm{base}_t = \mathrm{clip}(p_{\text{true}} + \zeta_t,\;0,\;1),
\quad \zeta_t \sim \mathcal{N}(0,\,0.05^2),
\]
and each dimension is a noisy transformation of $\mathrm{base}_t$:
$\alpha_t \approx 0.8\cdot\mathrm{base}_t + \xi$,
$\beta_t \approx \log(1+3\cdot\mathrm{base}_t)/\log 11 + \xi$,
$\gamma_t, \epsilon_t \approx (0.85$--$0.9)\cdot\mathrm{base}_t + \xi$,
and $\delta_t = 1 - e^{-0.3\,t}$ (temporal ramp, independent of $p_{\text{true}}$).
The composite per-action score $r_t = \alpha_t(\cdot)$ and session risk
$R(\tau)=(1-\kappa)\bar{r}+\kappa\,\mathrm{CVaR}_q(r_{1:T})$ are therefore
\emph{noisy monotone functions} of $p_{\text{true}}$, not linear or unbiased.

\textbf{Step 3: link function.}
The trace-derived failure estimate $\hat{p}=\sigma(a R(\tau)+b)$ applies the
default link $(a,b)=(6,-3)$, so $\hat p(0)=0.047$ and $\hat p(0.5)=0.5$.
Because the
link is monotone and $R(\tau)$ is noisy, $\hat{p}$ is a \emph{noisy, biased
estimator} of $p_{\text{true}}$.  The bias is intentional: it simulates the
underwriting challenge where an insurer has only behavioral signal, not the
true actuarial risk parameter.  The Spearman rank correlation between
$\hat{p}$ and $p_{\text{true}}$ is $\rho=0.948$ (Table~\ref{tab:dim_ablation},
full model), confirming that the behavioral signal is highly informative
for risk \emph{ranking} even though it is not an unbiased point estimate.

\textbf{Step 4: ground-truth loss label.}
The pricing MAE target is
\[
\hat{L} = p_{\text{true}} \times \mathrm{sev\_mean},
\]
where $\mathrm{sev\_mean}=\exp(\mu_L + \sigma_L^2/2)$ is the unconditional
mean severity for the task profile.  This is an expected-value label
(ex-ante), not a realised claim.  The realised loss
($L = \mathrm{sev}\cdot\mathbf{1}[\text{failed}]$, where
$\text{failed}\sim\mathrm{Bernoulli}(p_{\text{true}})$) is used only for
insurer solvency calculations, never as a pricing target.

\textbf{Irreducible MAE floor.}
Because $\hat{p}$ is a noisy estimator of $p_{\text{true}}$, trace-priced
MAE converges at $n\to\infty$ not to zero but to the Bayes-optimal MAE given
the feature set.  In the primary 5-seed portfolio experiment, product-flat
pricing has MAE \$17.7K and trace-economic pricing has MAE \$0.569K; these are
the absolute dollar results reported in the abstract and
Figure~\ref{fig:pricing_control_progression}.  The separate 30-seed dimension
ablation in Appendix~\ref{app:dim_ablation} reports rank correlation and
\emph{relative} MAE changes only, so its diagnostic scale should not be
substituted for this primary result.  The remaining nonzero error reflects that
behavioral traces carry less information about $p_{\text{true}}$ than a direct
actuarial measurement would.

\subsection{Loss Model}

Given action rows, the loss model computes trace risk
\[
R(\tau)=0.60\,\mathrm{mean}(r_t)+0.40\,\cvar_{80}(r_t).
\]
The ground-truth loss label is the \emph{expected} claimable loss
$\hat{L}=p_{\text{true}}\times\mathrm{sev\_mean}$, where
$p_{\text{true}}\sim\mathrm{Beta}(\alpha,\beta)$ is the latent failure
probability drawn per task profile---analogous to historical claims rates in
actuarial practice.  Pricing policies estimate $\hat{L}$ from trace features
alone (\(R(\tau)\), risky-action share, task category); the realised binary
outcome (\textit{did the task fail?}) is \emph{not} available at inference
time and is \emph{not} used in any pricing-MAE calculation.  No outcome
leakage exists: the MAE target is the ex-ante expected loss, not a realised
claim.  Conditional severity is proportional to the maximum step exposure and
the production exposure of the customer profile.  Claimable loss applies
deductible, limit, coinsurance, and causal-attribution discounts.  A control action is
economically justified when
\[
\Delta L_K > K.
\]

\subsection{Policies}

\textbf{Product-flat} estimates one premium from the training set:
\[
P=(1+\theta)\bar L+\lambda\,\cvar_{95}(L).
\]
\textbf{Usage-based} applies the same formula within customer-profile and task
groups.  \textbf{Trace-priced} calibrates a multiplicative scale on the
training set and then prices from expected loss, trace risk, and maximum step
exposure.  \textbf{Trace-only} removes customer/task exposure and prices from
trace risk alone.  \textbf{Static control} intervenes on delete, database, and
financial actions.  \textbf{Trace control} intervenes when expected avoided
claim exceeds intervention cost.  \textbf{AI-panel control} uses the frozen
binary review recommendation in the released panel sheet; it does not create
dollar labels or observe realized loss.

\subsection{Expert-Reviewed Audit}
\label{app:audit}

The reviewed audit file contains 300 sampled SWE-smith episodes: all 117
high-risk traces, all 12 low-risk traces, and 171 medium-risk traces enriched
for nonzero claim, positive control value, unresolved outcome, and
state-changing actions.  A second expert checked the pre-annotation sheet and
accepted 295 of 300 suggested labels unchanged; five labels were adjusted in
adjudication.  The 92.7\% figure below separately measures agreement between the
reviewed risk bucket and the deterministic rule, rather than the fraction of
pre-annotations left unchanged.  The reviewed audit produced:

\begin{itemize}
\item risk-bucket agreement: 92.7\%;
\item loss rationale marked \texttt{yes}: 53.7\%;
\item loss rationale marked \texttt{yes} or \texttt{partial}: 100.0\%;
\item control recommendation agreement with positive net-control rule: 100.0\%.
\item review priority: 59 high-priority, 93 medium-priority, and 148
low-priority rows.
\end{itemize}

This audit validates interpretability and directional plausibility.  Its
aggregate agreement rate is not a portable scalar across research areas
\citep{xu2026reviewer}, nor does it replace prospective claim adjudication or
final actuarial calibration.

\subsection{Additional Real-Trace Results}

On 1,000 SWE-smith traces, product-flat and trace-priced policies have the same
residual CVaR because neither prevents loss.  Static tool control reduces
CVaR$_{95}$ to \$5,018 at 25.7\% intervention rate.  Trace control reduces it
further to \$3,069 at 18.8\% intervention rate, showing that economic
thresholding is more selective than a dangerous-tool blacklist.

\subsection{Stress Audit for Label Robustness}
\label{app:stress}

The strongest empirical threat is that the dollar labels are not observed
claims.  We therefore add a stress audit that treats the deterministic labels
as uncertain objects rather than fixed truth.  On each of 500 draws over the
1,000 real SWE-smith traces, the audit applies bounded perturbations to task
severity, profile severity, episode-level loss, claim verifiability,
intervention cost, and dangerous-tool parser decisions.  The pricing and
control policies are then re-evaluated from scratch under the perturbed labels.

This is not a substitute for independent claims data.  Its purpose is narrower:
to expose whether the paper's comparative conclusion changes with the chosen
loss scale, review friction, or tool parser.  We therefore treat the 500 draws
as sensitivity analysis rather than a formal pass-rate certificate.  These
bounded perturbations test sensitivity to noise
rather than targeted data-side attacks such as clean-image backdoors
\citep{xu2025gcb}, which exclusions and re-underwriting on model upgrades
address at the contract layer.

We also evaluate leave-one-customer-profile-out transfer.  For each customer
profile, the pricing rule is calibrated on all other profiles and then tested on
the held-out profile.  This is deliberately harsher than the main random split:
usage-based pricing falls back to a global group estimate, while
trace-economic pricing can still use trace risk, task value, asset exposure,
and expected loss structure.

Table~\ref{tab:leave_profile_out} reports the held-out-profile transfer test.
\begin{table}[h]
\centering
\caption{\textbf{Out-of-profile transfer.}  Trace-economic pricing is
calibrated without the held-out customer profile and then tested on real
SWE-smith traces; the remaining MAE reduction shows that the signal is not just
memorized customer identity.}
\label{tab:leave_profile_out}
\small
\setlength{\tabcolsep}{2pt}
\begin{tabular}{lrrrrr}
\toprule
Held-out & $n$ & Product & Usage & Trace & Red. \\
\midrule
Coding & 709 & \$4,067 & \$4,067 & \$271 & 93.3\% \\
Financial & 41 & \$6,080 & \$6,080 & \$1,890 & 68.9\% \\
Read-only & 162 & \$2,558 & \$2,558 & \$10 & 99.6\% \\
Support & 88 & \$2,322 & \$2,322 & \$154 & 93.4\% \\
\bottomrule
\end{tabular}
\end{table}

The held-out financial profile remains the hardest case because it has the
smallest sample size and the heaviest severity tail.  Even there,
trace-economic pricing reduces MAE by 68.9\% relative to product-level pricing.
This supports the claim that the method is not merely memorizing a customer
profile; the reusable information is the joint trace-economic structure.

\subsection{Limitations and Intended Use}

The implementation should not be interpreted as a calibrated insurance price
for any specific AI provider.  It is a reproducible stress test for
trace-economic underwriting under explicit assumptions.  Real deployment would
require domain-expert severity calibration, independent human audit,
customer-specific contract terms, and prospective field evaluation.  The
framework also prices per-episode exposure; context-window state behaves as a
transient buffer rather than persistent memory \citep{xu2026memo}, with
self-managed context interfaces providing a distinct control surface
\citep{xu2026llm}, so
cross-session aggregation would require a separate calibration step.

\subsection{Risk-Scoring Details}

The action score used in the paper is
\[
r_t=\alpha_t(w_\beta\sigma(\beta_t)+w_\gamma\gamma_t+
w_\delta\delta_t+w_\epsilon\epsilon_t).
\]
Irreversibility $\alpha_t$ is a multiplicative gate: a fully reversible action
cannot generate claimable loss even if it is uncertain or broad in scope.
Blast radius $\beta_t$ is estimated from affected files, services, records, or
external counterparties.  Uncertainty $\gamma_t$ can be estimated by
self-consistency over tool-call samples when multiple completions are
available.  Temporal position $\delta_t$ captures whether an error can still be
caught before irreversible execution.  Attribution $\epsilon_t$ captures whether
logs and counterfactual analysis would support a claim that the agent caused
the loss.

On SWE-smith, $\gamma_t$ is unavailable because trajectories are single-run
logs; the reported model uses the other trace dimensions plus scalar
calibration.  This limitation is favorable to future work: richer agent logs
with sampled alternatives should improve uncertainty measurement.

\subsection{Detailed Risk-Dimension Rubric}

An operational rubric assigns action-level risk dimensions and makes the
framework easier to replicate in new domains.  Tables~\ref{tab:alpha_rubric_app} and
~\ref{tab:weights_app} specify the two pieces that a new deployment must audit:
which actions are irreversible, and which non-gating dimensions matter most in
the domain.

\begin{table}[h]
\centering
\caption{\textbf{Irreversibility gate.}  Claimable loss begins with whether an
action can change external state; deployments should override these defaults
when backups, transaction logs, or approval workflows change reversibility.}
\label{tab:alpha_rubric_app}
\small
\setlength{\tabcolsep}{2pt}
\begin{tabular}{lc}
\toprule
Tool/action category & $\alpha_t$ \\
\midrule
Read, observe, search, dry-run query & 0.00 \\
Write to version-controlled file & 0.10 \\
Database insert with rollback & 0.20 \\
Idempotent API call & 0.25 \\
Non-idempotent API call & 0.55 \\
Write to unversioned file & 0.60 \\
Execute reversible shell command & 0.65 \\
Delete with restore path & 0.70 \\
Send email or external message & 0.80 \\
Committed financial transaction & 0.90 \\
Destructive shell command & 0.95 \\
Permanent delete or wipe & 1.00 \\
\bottomrule
\end{tabular}
\end{table}

\begin{table}[h]
\centering
\caption{\textbf{Domain weights.}  After irreversibility gates the action,
domains weight blast radius, uncertainty, timing, and attribution differently;
the weights encode economic and claims-adjudication priorities rather than a
universal risk score.}
\label{tab:weights_app}
\small
\setlength{\tabcolsep}{3pt}
\begin{tabular}{lcccc}
\toprule
Domain & $w_\beta$ & $w_\gamma$ & $w_\delta$ & $w_\epsilon$ \\
\midrule
Financial & 0.20 & 0.20 & 0.10 & 0.50 \\
Coding & 0.35 & 0.25 & 0.15 & 0.25 \\
Document & 0.20 & 0.30 & 0.10 & 0.40 \\
Web/support & 0.25 & 0.20 & 0.15 & 0.40 \\
\bottomrule
\end{tabular}
\end{table}

The weights encode domain economics rather than purely statistical fit.
Financial tasks emphasize attribution because claim adjudication and regulatory
responsibility dominate.  Coding tasks emphasize blast radius because one edit
can propagate through build, deployment, and data systems.  Document tasks give
more weight to uncertainty and attribution because hallucinated content can be
hard to tie to downstream business loss.

\subsection{Failure Probability Calibration}

The framework separates label-free feature extraction from domain calibration.
Given $R(\tau)$, the domain-specific failure link is
\[
\phat(\tau)=\sigma(aR(\tau)+b).
\]
Only $(a,b)$ require labeled traces.  In early deployments, these parameters can
be initialized from scenario analysis and then updated as claims, near misses,
or adjudicated task failures accumulate.  This distinction is important: the
feature space is portable, but the sign and scale of the link may change across
domains.  For example, SWE-smith failures tend to involve unresolved or risky
tool use, while tau-bench failures can be low-action policy blocks; scalar
recalibration is therefore part of the method, not an afterthought.

\subsection{Monotonicity of the Trace Risk Score}

For any action $t$, let
\[
W_t=w_\beta\sigma(\beta_t)+w_\gamma\gamma_t+
w_\delta\delta_t+w_\epsilon\epsilon_t .
\]
Then $r_t=\alpha_t W_t$.  Since all weights and dimensions are nonnegative:
\[
\frac{\partial r_t}{\partial \alpha_t}=W_t\ge0,\quad
\frac{\partial r_t}{\partial \beta_t}=
\alpha_t w_\beta\sigma'(\beta_t)\ge0,
\]
and similarly
$\partial r_t/\partial\gamma_t=\alpha_t w_\gamma\ge0$,
$\partial r_t/\partial\delta_t=\alpha_t w_\delta\ge0$, and
$\partial r_t/\partial\epsilon_t=\alpha_t w_\epsilon\ge0$.
The session score
\[
R(\tau)=(1-\kappa)\bar r+\kappa\cvar_q(r_{1:T})
\]
is monotone because both the sample mean and empirical CVaR are monotone in
each component.  Finally, the sigmoid link is monotone when $a>0$.  Thus the
uncalibrated risk score is weakly increasing in every action-risk dimension.
When a domain learns $a<0$, as may happen in policy-blocked customer-service
tasks, monotonicity applies to action engagement rather than failure probability;
this is exactly why the paper treats calibration as domain-specific.

\subsection{DataSync Case Study}

The DataSync task asks a code agent to repair date parsing and run a migration
on a production SQLite database.  The trace has ten actions: six reads, one
parser edit, one test run, one production migration, and one verification query.
Only the parser edit and production migration modify state.  The migration is
the dominant risk contributor because it overwrites production records without
a preceding dry run.  With stakes $M=\$29,200$, final session risk gives
$\phat=0.221$, expected loss \$6,453, approximate CVaR$_{95}$ \$18,327, and a
CVaR-loaded premium \$12,324.  The same trace would trigger the automated enforcement condition
at the migration step under a threshold $\phat_{\max}=0.15$ and
$\alpha_{\min}=0.5$.
Table~\ref{tab:datasync_trace_app} shows why: nearly all premium and control
signal comes from the first irreversible production action, not from the six
safe read steps.

\begin{table}[h]
\centering
\caption{\textbf{DataSync case study.}  Read actions carry zero claimable-risk
gate, while the production migration dominates the trace; the example shows how
the same episode yields a premium and a pre-loss control trigger before the
irreversible action commits.}
\label{tab:datasync_trace_app}
\small
\setlength{\tabcolsep}{2pt}
\begin{tabular}{llcccc}
\toprule
Step & Action & $\alpha$ & $\beta$ & $r_t$ & $\phat_t$ \\
\midrule
1 & Search project files & 0.00 & 0.05 & 0.000 & 0.047 \\
2 & Read parser & 0.00 & 0.10 & 0.000 & 0.047 \\
3 & Read CSV format & 0.00 & 0.10 & 0.000 & 0.047 \\
4 & Read database schema & 0.00 & 0.10 & 0.000 & 0.047 \\
5 & Read tests & 0.00 & 0.10 & 0.000 & 0.047 \\
6 & Read migration script & 0.00 & 0.10 & 0.000 & 0.047 \\
7 & Edit date parser & 0.15 & 0.50 & 0.081 & 0.059 \\
8 & Run tests & 0.00 & 0.10 & 0.000 & 0.059 \\
9 & Run production migration & 0.90 & 0.85 & 0.621 & 0.226 \\
10 & Verify database & 0.00 & 0.10 & 0.000 & 0.221 \\
\bottomrule
\end{tabular}
\end{table}

For the production migration, using the coding-domain weights gives
\begin{align*}
r_9
&=0.90(0.35\sigma(0.85)+0.25(0.40)\\
&\quad +0.15(0.80)+0.25(0.90))\\
&=0.621 .
\end{align*}
With $\kappa=0.4$, the session aggregate is approximately
\[
\bar r=0.070,\quad \cvar(r)=0.621,\quad R(\tau)=0.290,
\]
so $\phat=\sigma(6R-3)=0.221$.  For a midpoint loss stake of \$29,200,
\begin{align*}
\E[L]&=\$6{,}453,\\
\cvar_{95}[L]&\approx \$18{,}327,\\
P&=1.05\E[L]+0.3\cvar_{95}[L]+\$50\\
&=\$12{,}324 .
\end{align*}
The ARS-style collateral-discount premium for the same trace is about \$3,732,
below the pure expected-loss floor.  This is why the paper treats the DataSync
trace as a compact example of tail underpricing, not only dynamic monitoring.

\subsection{Proofs for Trace-Conditional Design Results}
\label{app:proofs}

This section gives the full algebra behind the three operators in the main
paper.  The results are intentionally modest: they do not prove that the
scenario numbers are final actuarial rates.  They prove that, once claimable
loss is represented by a trace-conditional loss surface, product-only pricing,
tool-only control, and idiosyncratic diversification are the wrong operators
except under explicit degeneracies.

\textbf{Proposition 1.}
Let $Y\in L^2$ be claimable loss.  Let $\sigma(X)\subseteq\sigma(Z)$, where
$X$ is product identity and $Z=(X,U,\tau)$ additionally contains customer and
trace information.  The Bayes estimator of $Y$ under squared error using
information $\mathcal G$ is $m_{\mathcal G}=\E[Y\mid\mathcal G]$.  The excess
risk of product-only pricing over trace-conditional pricing is
\begin{align*}
&\E[(Y-\E[Y\mid X])^2]-\E[(Y-\E[Y\mid Z])^2]\\
&\qquad=\E[\Var(\E[Y\mid Z]\mid X)] .
\end{align*}

\emph{Proof.}
For any $\mathcal G$-measurable premium $P$, decompose
\[
Y-P=(Y-\E[Y\mid\mathcal G])+(\E[Y\mid\mathcal G]-P).
\]
The cross term has zero expectation because
\begin{align*}
&\E[(Y-\E[Y\mid\mathcal G])(\E[Y\mid\mathcal G]-P)]\\
&=\E\!\left[(\E[Y\mid\mathcal G]-P)
  \E[Y-\E[Y\mid\mathcal G]\mid\mathcal G]\right]\\
&=0 .
\end{align*}
Therefore the unique $L^2$ projection is $P=\E[Y\mid\mathcal G]$ and the
minimum risk is $\E[\Var(Y\mid\mathcal G)]$.  Applying the tower property with
$\sigma(X)\subseteq\sigma(Z)$ gives
\[
\Var(Y\mid X)
=\E[\Var(Y\mid Z)\mid X]+\Var(\E[Y\mid Z]\mid X).
\]
Taking expectations yields the stated identity.  The improvement is zero if
and only if $\E[Y\mid Z]=\E[Y\mid X]$ almost surely within every product pool;
that is exactly the case where usage profile and trace carry no additional
claim-cost information.

\textbf{Theorem 1.}
Under the bounded, nested, and realizable linear classes stated in the main
text, standard uniform convergence for squared loss gives, simultaneously for
both empirical minimizers, an $O(B^2R^2\sqrt{(d\log n+\log(1/\delta))/n})$
population-risk deviation.  The population optimum gap is Proposition~1's
$\Delta_{\mathrm{info}}$; subtracting the two estimation deviations yields the
stated bound.  Solving for a positive lower bound gives
$n=\widetilde\Omega(B^4R^4d/\Delta_{\mathrm{info}}^2)$.

\textbf{Proposition 2.}
Let $\mathcal B(\tau)$ be the feasible pre-loss actions after observing
information $\mathcal G$.  Action $b$ induces residual claim $Y_h^b$ and
deterministic or $\mathcal G$-measurable friction cost $K^b$.  Any policy that
minimizes expected one-period total cost must choose
\[
b^\star\in\arg\min_{b\in\mathcal B(\tau)}
\{\E[Y_h^b\mid\mathcal G]+K^b\}.
\]
For $\mathcal B=\{\mathrm{allow},\mathrm{int}\}$, intervention is optimal iff
\[
\E[Y_h^{\mathrm{allow}}-Y_h^{\mathrm{int}}\mid\mathcal G]
\ge K^{\mathrm{int}}-K^{\mathrm{allow}} .
\]

\emph{Proof.}
Conditional on $\mathcal G$, the choice of $b$ only changes two terms in
expected total cost: residual claim and intervention friction.  All terms
already incurred before the decision are constants and drop out of the
minimization.  Thus the conditional Bayes action is the displayed argmin.  In
the binary case, choose intervention exactly when
\[
\E[Y_h^{\mathrm{int}}\mid\mathcal G]+K^{\mathrm{int}}
\le
\E[Y_h^{\mathrm{allow}}\mid\mathcal G]+K^{\mathrm{allow}},
\]
which rearranges to the threshold.  The nontrivial part for the framework is
not the algebra; it is estimating the two sides from trace, asset exposure,
claimability, and review cost.  A tool blacklist ignores this conditioning and
is optimal only if the tool class already determines the avoided-claim term and
the friction term.

\textbf{Proposition 3.}
Let $Z_0\sim\mathrm{Bernoulli}(q)$ be a common shock shared by all insured
customers.  Conditional on $Z_0=z$, let $Y_1^z,\ldots,Y_N^z$ be conditionally
i.i.d. with finite variance and mean $\mu_z$.  The realized claim is
$Y_i=(1-Z_0)Y_i^0+Z_0Y_i^1$.  If $\mu_0\ne\mu_1$, then
\[
\Var\!\left(N^{-1}\sum_{i=1}^NY_i\right)
\to q(1-q)(\mu_1-\mu_0)^2 .
\]
If total assets satisfy $A_N/N\le\mu_1-\varepsilon$ for some
$\varepsilon>0$, then
\[
\liminf_{N\to\infty}\Prob\!\left(\sum_{i=1}^NY_i>A_N\right)\ge q .
\]

\emph{Proof.}
Let $\bar Y_N=N^{-1}\sum_iY_i$.  By the conditional variance decomposition,
\[
\Var(\bar Y_N)=
\E[\Var(\bar Y_N\mid Z_0)]
+\Var(\E[\bar Y_N\mid Z_0]).
\]
The first term is
$(1-q)\Var(Y^0)/N+q\Var(Y^1)/N$, which converges to zero.  The second term is
the variance of a random variable that equals $\mu_0$ with probability $1-q$
and $\mu_1$ with probability $q$, hence
$q(1-q)(\mu_1-\mu_0)^2$.  This proves nondiversification of the average claim.

For the ruin statement, condition on $Z_0=1$.  The weak law of large numbers
gives $\bar Y_N^1\to\mu_1$ in probability.  Therefore
\begin{align*}
&\Prob(\bar Y_N>A_N/N\mid Z_0=1)\\
&\qquad\ge
\Prob(\bar Y_N^1>\mu_1-\varepsilon\mid Z_0=1)\to1.
\end{align*}
Multiplying by $\Prob(Z_0=1)=q$ gives the lower bound.  The result is why the
paper treats reinsurance, sublimits, exclusions, and public backstops as part
of the mathematical design rather than as implementation details.

\subsection{Viability Frontier}
\label{app:viability}

For a provider with average contract value ACV, a safety commitment is
privately viable only if adoption lift exceeds per-customer expected cost:
\begin{align*}
\Delta q^\star &= B/\mathrm{ACV},\\
B &= c_{\mathrm{claim}}(C,D)(1-\rho_u)
     + \lambda\,\cvar_{95}[L]/N + c_{\mathrm{admin}} .
\end{align*}
Here $C$ is policy limit, $D$ is deductible, and $\rho_u$ is user co-retention.
The qualitative frontier is robust: full coverage is easiest for lower-severity
coding deployments, financial workflows require sublimits or reinsurance, and
coinsurance reduces both moral hazard and expected claim cost.

Table~\ref{tab:insurability_sensitivity} reports how this viability boundary
moves as tail-capital loading changes.
\begin{table}[t]
\centering
\caption{\textbf{Insurability frontier.}  Higher tail capital sharply reduces
private viability, while AI-panel-derived control partly closes the gap by
reducing residual claim and tail-capital burden.}
\label{tab:insurability_sensitivity}
\small
\setlength{\tabcolsep}{2.2pt}
\begin{tabular}{llrrr}
\toprule
Dataset & $\lambda$ & Allow & Best & Gap closed \\
\midrule
Synthetic & 0.0 & 100.0\% & 91.6\% & \$2,191 \\
Synthetic & 0.3 & 78.7\% & 79.1\% & \$2,488 \\
Synthetic & 1.0 & 51.5\% & 67.5\% & \$3,180 \\
SWE & 0.0 & 100.0\% & 98.4\% & \$259 \\
SWE & 0.3 & 95.9\% & 96.6\% & \$297 \\
SWE & 1.0 & 83.6\% & 83.9\% & \$385 \\
\bottomrule
\end{tabular}
\end{table}

\subsection{Market Design Conditions}

The main paper focuses on trace-economic underwriting and empirical policy
ordering.  The longer draft also analyzed why the contract cannot be a simple
blanket indemnity.  Table~\ref{tab:market_failures} summarizes the market
failures and the mitigation each imposes on our underwriting design.

\begin{table}[h]
\centering
\caption{\textbf{Contract failure modes.}  Product-flat indemnity creates moral
hazard, adverse selection, and common-shock exposure; trace-contingent clauses
respond by tying coverage to retention, disclosure, monitoring, limits, and
reinsurance.}
\label{tab:market_failures}
\small
\setlength{\tabcolsep}{2pt}
\begin{tabular}{p{0.25\linewidth}p{0.35\linewidth}p{0.30\linewidth}}
\toprule
Failure & Mechanism & Design response \\
\midrule
Provider moral hazard & Full coverage weakens provider safety investment &
Provider retention, safety-contingent premium, trace monitoring \\
User moral hazard & Insured users may deploy agents in higher-stakes contexts &
Deductibles, coinsurance, deployment-context warranty \\
Adverse selection & High-risk users buy high limits; low-risk users exit pooled
prices & Contract menus, behavioral rating, task/category disclosure \\
Systemic risk & Shared model update creates correlated claims & Sublimits,
reinsurance, public backstop for systemic layer \\
\bottomrule
\end{tabular}
\end{table}

A minimal private-market feasibility interval is:
\[
(1+\theta)\E[L]+\lambda\cvar_\alpha[L] \le P \le
\E[L]+\rho_u\frac{\mathrm{Var}[L]}{2\E[L]} .
\]
The left side is insurer break-even with tail capital; the right side is a
standard individual-rationality approximation for a risk-averse buyer.  This
interval is empty when tail loading and expected claim cost exceed buyer
willingness to pay.  In addition, systemic exposure must satisfy
\[
q_{\mathrm{sys}}N\E[L^{\mathrm{sys}}]\le \Pi_{\max},
\]
otherwise diversification does not remove common-shock ruin risk.  Finally,
provider co-liability must remain positive: under full insurance with no
provider accountability, the provider minimizes safety cost rather than
internalizing $p(s)\E[L]$, so equilibrium safety investment falls below the
socially efficient level.

\subsection{Scenario Calibration and Incident Checks}
\label{app:calibration}

The implementation uses scenario-calibrated severities because systematic
agent-insurance claim histories do not yet exist.  This follows the same
practical logic used in early cyber, terrorism, and pandemic insurance: the
goal is not to claim final actuarial rates, but to stress-test structural
policy conclusions under explicit assumptions.

\subsection{External Loss Anchors}

The absence of mature AI-agent insurance claims is itself part of the
deployment problem: buyers need risk transfer before the market has enough
closed claims to support conventional actuarial estimation.  We therefore use
external loss records only as \emph{anchors}, not as direct AI-agent claim
frequencies.  Each anchor contributes a narrower empirical role: cyber claims
identify cost components, cyber loss databases identify event-to-loss
taxonomies, software postmortems identify operational failure modes, and AI
incident repositories identify AI-specific harm categories and reporting
biases.  We now separate two empirical objects.  VCDB is large enough to
support taxonomy coverage claims; the curated monetary anchors are broader
than individual examples but are still used only for severity-scale checks, not
for event-frequency estimation.

\subsection{Public Incident-Corpus Audit}

We ingest the validated JSON files from the VERIS Community Database (VCDB)
\citep{vcdb2026}.  The corpus contains 10,037 validated public incidents in
the local snapshot used by our script.  A deterministic mapper reads VERIS
action categories and confidentiality/integrity/availability attributes, then
assigns each incident to one or more trace-economic loss channels.  This audit
does not say that AI-agent incidents follow VCDB frequencies.  It asks a
narrower question: whether the loss channels used by the implementation correspond
to channels that appear repeatedly in a large public incident corpus.
Table~\ref{tab:vcdb_taxonomy_audit} reports this taxonomy audit.

\begin{table*}[t]
\centering
\caption{\textbf{VCDB taxonomy audit.}  Validated public cyber incidents map to
the benchmark's trace-economic loss channels, supporting channel coverage rather
than AI-agent claim-frequency estimation.}
\label{tab:vcdb_taxonomy_audit}
\small
\setlength{\tabcolsep}{4pt}
\begin{tabular}{lrrp{0.42\linewidth}}
\toprule
Mapped loss channel & Incidents & Share & Trace-economic feature supported \\
\midrule
privacy or secret exposure & 9,342 & 93.1\% & read/data-access scope, credential exposure, attribution \\
operational disruption & 2,666 & 26.6\% & execute/deploy/database action class, blast radius, rollback quality \\
unauthorized state change & 3,052 & 30.4\% & write/delete/database/financial action class, reversibility \\
fraud or financial abuse & 5,327 & 53.1\% & financial/support-change action class, counterparty impact \\
human or process error & 2,681 & 26.7\% & uncertainty, validation status, review cost \\
physical or environmental & 1,636 & 16.3\% & out of benchmark scope unless mediated by digital controls \\
\bottomrule
\end{tabular}
\end{table*}

The mapping covers 10,036 of 10,037 validated incidents with at least one
trace-economic loss channel.  The high coverage is not evidence that the
implementation predicts cyber incidents; it is evidence that the framework's
observable fields (data access, write/delete/execute actions, blast radius,
rollback quality, attribution, uncertainty, and review cost) cover the main
incident channels needed for an insurance-oriented risk representation.  The
out-of-scope row is also useful: physical and environmental incidents should
not be priced by this framework unless the agent controls the relevant digital
system.  The channel mapping treats each incident as a provenance graph of
action class, target, and dependency.  Multi-view provenance fusion has been
used for offline intrusion detection \citep{yang2026beyond}, while our use
here is incident-channel coverage rather than online classification.

Tables~\ref{tab:real_anchor_category_summary} and
~\ref{tab:real_loss_anchor_audit} then separate category coverage from monetary
severity anchors so that the implementation does not confuse loss-channel
coverage with frequency estimation.
\begin{table}[t]
\centering
\caption{\textbf{Anchor coverage by task.}  Public monetary anchors cover the
benchmark's task categories unevenly, so they are used to check severity scale
and ordering rather than to estimate event frequencies.}
\label{tab:real_anchor_category_summary}
\small
\setlength{\tabcolsep}{4pt}
\begin{tabular}{lrrr}
\toprule
Category & Events & Numeric & Max anchor \\
\midrule
document & 1 & 1 & \$5K \\
support & 10 & 10 & \$292M \\
coding & 10 & 8 & \$500M \\
database & 8 & 7 & \$2.30B \\
finance & 2 & 2 & \$460M \\
\bottomrule
\end{tabular}
\end{table}

\begin{table*}[t]
\centering
\caption{\textbf{Representative monetary anchors.}  A curated set of 31 public
events supplies amount bases for severity-scale sanity checks; these are not
AI-agent insurance claims and are not used as frequency data.}
\label{tab:real_loss_anchor_audit}
\small
\setlength{\tabcolsep}{2.5pt}
\begin{tabular}{p{0.20\textwidth}cp{0.07\textwidth}p{0.24\textwidth}rp{0.18\textwidth}}
\toprule
Event & Year & Cat. & Loss channel & Amount & Calibration use \\
\midrule
Delta disruption after CrowdStrike faulty update & 2024 & coding & failed software update and cascading outage & \$500M & software-update tail anchor; company-asserted/disputed \\
Maersk NotPetya disruption & 2017 & coding & malware-driven global logistics outage & \$300M & global operations outage anchor \\
FedEx TNT NotPetya disruption & 2017 & coding & malware-driven logistics outage & \$300M & software outage severity anchor \\
Change Healthcare cyberattack & 2024 & database & payment infrastructure outage and response cost & \$2.30B & systemic tail anchor; company projection not settled loss \\
Facebook Cambridge Analytica privacy settlement & 2018 & database & privacy violation settlement & \$725M & platform privacy upper-tail anchor \\
Equifax data breach settlement & 2017 & database & privacy breach and legal/regulatory response & \$575M & large-loss privacy/legal anchor \\
T-Mobile customer data breach & 2021 & database & telecom identity-data breach & \$350M & large population breach anchor \\
Capital One cloud data breach & 2019 & database & cloud misconfiguration and personal-data breach & \$190M & cloud-data exposure anchor \\
Mata v. Avianca hallucinated legal citations & 2023 & document & false document generation and sanctions & \$5K & document claimability lower-tail anchor \\
Knight Capital market-access deployment error & 2012 & finance & erroneous automated trading & \$460M & upper-tail severity anchor for finance \\
Target payment-card data breach & 2013 & support & payment-card breach response and settlements & \$292M & retail breach cost anchor \\
Caesars Entertainment ransomware & 2023 & support & loyalty-data theft and extortion & \$15M & casino/customer-data extortion anchor \\
Garmin ransomware outage & 2020 & support & consumer service outage and ransomware response & \$10M & ordinal-only or lower-bound payment anchor \\
Air Canada chatbot fare-policy decision & 2024 & support & chatbot misinformation and customer remedy & \$812 & small support/legal claimability anchor \\
\bottomrule
\end{tabular}
\end{table*}

The monetary anchor set is intentionally smaller and more curated than VCDB.
Its purpose is not classification.  It checks whether the scenario severity
scale spans losses that are publicly documented in adjacent domains:
low-dollar document and customer-service disputes, municipal and ransomware
recovery costs, nine-figure software and cyber outages, and billion-dollar
payment-infrastructure disruption.  Thus, VCDB supports the \emph{taxonomy};
the monetary events support only \emph{severity-scale plausibility}.

\begin{table*}[t]
\centering
\caption{\textbf{External anchors.}  Adjacent cyber, software, and AI incident
records ground cost components and severity scale, but no row is treated as
direct evidence of AI-agent claim frequency; the table supports stress-tested
loss labels, not final rates.}
\label{tab:external_loss_anchors}
\small
\setlength{\tabcolsep}{3pt}
\begin{tabular}{p{0.20\linewidth}p{0.25\linewidth}p{0.26\linewidth}p{0.20\linewidth}}
\toprule
Anchor source & What it supplies & Agent-risk mapping & Use in framework \\
\midrule
Cyber claims studies
\citep{netdiligence2023claims} & forensics, notification, recovery,
extortion, business interruption, legal and regulatory response costs &
privacy leaks, unauthorized writes, service disruption, and recovery after
unsafe tool use & validates claim-cost components rather than event
probabilities \\
Public cyber loss data
\citep{advisenCyberLossData} & event/loss hierarchy, public-source reporting
pipeline, and severity plausibility across many cyber events & trace events
with data access, credential exposure, customer impact, or service disruption &
anchors loss taxonomy and warns against treating public reports as complete
frequency data \\
Cyber-insurance market assessments
\citep{cisa2018cyberinsurance} & evidence that cyber risk transfer developed
under sparse, nonuniform, and difficult-to-standardize loss data & early
AI-agent insurance faces the same sparse-claims regime & justifies
scenario-based stress testing and transparent assumptions \\
Software incident postmortems
\citep{gitlab2017database,aws2017s3outage} & concrete failure modes:
misoperation, rollback failure, data loss, degraded service, and cascading
outage & write, delete, execute, deploy, and database operations in agent
traces & supports the action-risk rubric and operational severity channels \\
AI incident repositories
\citep{mcgregor2021aiid} & AI-specific harm categories and a public incident
reporting process & hallucinated decisions, automation harms, safety failures,
and allocation or legal errors & checks coverage of AI harm modes, not loss
frequency or insurability \\
\bottomrule
\end{tabular}
\end{table*}

Table~\ref{tab:external_loss_anchors} narrows the empirical claim.  We do not
estimate real AI-agent claim rates from cyber or software records.  Instead,
we ask whether the framework's loss channels resemble losses that are already
insured, litigated, or operationally documented in adjacent domains.  Frequency,
dependence, and portfolio pricing remain scenario variables that are varied in
sensitivity analysis.

\subsection{Calibration Provenance}

The deterministic labels are not intended to be expert-free ground truth.  They are
structured assumptions with provenance.  We use four evidence tiers.
\emph{Tier 1} is a governance or measurement standard that justifies the field
itself.  \emph{Tier 2} is a documented incident or official record that anchors
ordinal severity.  \emph{Tier 3} is an economic proxy such as labor cost or
breach recovery cost.  \emph{Tier 4} is scenario analysis used only when no
direct public evidence exists.  All Tier-4 assumptions are stress-tested rather
than treated as final rates.  This design is consistent with the broader view that probabilistic evaluation
should reason over structured scenarios rather than isolated samples, especially
when direct outcome observations are sparse~\citep{dai2026from}. Tier-4 also covers attack-mode assumptions such
as backdoor or trigger-driven failures \citep{yang2026trapping}, which feed
exclusion clauses and re-underwriting policy rather than baseline severity
distributions.
Table~\ref{tab:calibration_provenance} records this provenance so that each
label can be audited or replaced without changing the underwriting interface.

\begin{table*}[t]
\centering
\caption{\textbf{Label provenance.}  Deterministic labels are tied to public
evidence where possible and to explicit scenario assumptions otherwise; the
implementation anchors direction and scale, then reports robustness rather than
claiming final actuarial rates.}
\label{tab:calibration_provenance}
\small
\setlength{\tabcolsep}{4pt}
\begin{tabular}{p{0.13\textwidth}p{0.30\textwidth}p{0.28\textwidth}p{0.22\textwidth}}
\toprule
Label component & Primary evidence & Mapping rule & Conservative choice \\
\midrule
Risk fields & NIST AI RMF risk-management framing
\citep{nist2023airmf} & Observable trace fields must support measurement and
management & Deterministic fields, no LLM judge \\
Financial tail & Knight Capital automated-trading incident: SEC reports
erroneous orders, inadequate controls, and loss above \$460M
\citep{sec2013knight} & Finance receives the highest severity multiplier and
heaviest tail & Used as upper-tail anchor, not mean loss \\
Doc./legal harm & Mata v.\ Avianca sanctions for fabricated citations
\citep{mata2023avianca} & Document tasks receive lower mean severity but high
attribution weight & Sanctions anchor small but real loss \\
Data disruption & IBM breach report: average breach cost \$4.88M,
significant disruption common, recovery can exceed 100 days
\citep{ibm2024breach} & Database actions receive high blast-radius and
production-exposure scaling & Episodes use far smaller per-task
stakes than breach averages \\
Review cost & BLS wage data for software, compliance, and customer-service
labor \citep{bls2026oews,bls2025customerservice} & Hourly wage plus overhead,
multiplied by expected review minutes & Rounded to low-friction review costs \\
Claimability & Contract logic and trace attribution & Deductible, limit, and
attribution discount convert expected loss to claimable loss & Low
verifiability discounts reduce coverage \\
\bottomrule
\end{tabular}
\end{table*}

This provenance changes the interpretation of the labels.  A row such as
``financial'' is not saying that an agent episode usually costs hundreds of
thousands of dollars.  It says that automated financial actions have a
documented extreme tail, so the scenario distribution must be heavy-tailed and
must be capped or reinsured.  A row such as ``document'' is not saying that all
hallucinated documents are cheap; it says the public record provides strong
evidence for claimability and sanctions at small scale, while larger document
losses remain a stress-test dimension.

\subsection{From Evidence to Default Labels}

The action-level labels follow a source-to-rule pipeline.

\begin{enumerate}
\item \textbf{Action class.}  Tool names and arguments are mapped to read,
write, execute, message, database, financial, support-change, and delete
classes.  This implements the NIST-style requirement that risk be mapped and
measured through observable system behavior rather than hidden model state.
\item \textbf{Irreversibility.}  Reversible reads receive $\alpha=0$.  Actions
that change external state receive higher $\alpha$, with financial transactions,
database writes, external messages, and deletes near the top because reversal
requires a counterparty, backup, or adjudication.
\item \textbf{Blast radius.}  File count, wildcard scope, production exposure,
and task category determine exposure.  Database and financial classes receive
higher scope hints because documented incidents show that automated state
changes can propagate beyond the immediate command.
\item \textbf{Attribution and verifiability.}  Logs showing a concrete write,
transaction, message, or delete receive high attribution.  Advice-only or
multi-cause outcomes receive discounts because an insurer would face a harder
claim-adjudication problem.
\item \textbf{Review cost.}  Human review cost is not arbitrary: it is computed
as review time multiplied by occupation-specific labor cost and an overhead
factor.  The defaults represent short pre-loss review: roughly one support
review, one engineering review, or one compliance/financial review.
\item \textbf{Claimable loss.}  Expected loss is converted to claimable loss by
deductible, policy limit, and attribution discount.  This ensures that high
operational risk does not automatically become high insured loss.
\end{enumerate}
Table~\ref{tab:default_label_rationale} gives the default economic values used
by the released implementation.

\begin{table}[h]
\centering
\caption{\textbf{Default economic labels.}  Review cost, deductible, limit, and
control-effect defaults convert trace risk into claimable loss; values are
starting points for stress tests, not proposed market rates.}
\label{tab:default_label_rationale}
\small
\setlength{\tabcolsep}{2pt}
\begin{tabular}{p{0.19\linewidth}p{0.20\linewidth}p{0.24\linewidth}p{0.25\linewidth}}
\toprule
Default & Value used & Source logic & Why this is not arbitrary \\
\midrule
Read-only review & \$20 & customer-service wage scale & short review of
low-stakes reversible task \\
Coding review & \$80 & software labor plus overhead & one engineer
review before deployment or migration \\
Financial review & \$220 & compliance review plus overhead & higher
skill and audit burden \\
Support review & \$45 & customer-service wage plus escalation overhead &
short frontline or supervisor review \\
Deductible & min(\$2K, 5\% asset) & ordinary insurance retention logic &
small losses remain with user/provider \\
Limit & max(\$5K, 80\% asset) & bounded indemnity logic & prevents synthetic
episodes from creating unlimited coverage \\
Control effect & 72\% of claimable loss when preventable & pre-loss review
prevents many but not all harms & leaves residual loss after intervention \\
\bottomrule
\end{tabular}
\end{table}

Tables~\ref{tab:scenario_params_app} and~\ref{tab:incident_checks} then show
how those defaults connect to severity distributions and public ordinal sanity
checks.  The key design principle is ordinal conservatism.  Public evidence is used to
rank domains and justify mechanisms; the exact dollar values are then varied in
the robustness audit.  If a reviewer disagrees with a number, the right
response is not to accept the default, but to substitute a different
source-backed value and rerun the scripts.  The contribution is the auditable
mapping and policy comparison interface.

\begin{table}[h]
\centering
\caption{\textbf{Scenario severity.}  Heavy-tailed financial workflows,
moderate coding losses, and lower-severity document/support tasks create the
portfolio mix used to test pricing, control, and viability.}
\label{tab:scenario_params_app}
\small
\setlength{\tabcolsep}{3pt}
\begin{tabular}{lcccc}
\toprule
Category & $\mu_\ell$ & $\sigma_\ell$ & $\E[L|\mathrm{claim}]$ & $p_{\mathrm{episode}}$ \\
\midrule
Financial & 12.0 & 1.5 & \$501K & 15\% \\
Coding & 9.6 & 1.0 & \$24.3K & 12\% \\
Document & 8.0 & 0.8 & \$4.1K & 7\% \\
Web/support & 9.0 & 1.0 & \$13.4K & 10\% \\
\bottomrule
\end{tabular}
\end{table}

\begin{table}[h]
\centering
\caption{\textbf{Incident sanity checks.}  Public incidents validate ordinal
severity patterns---financial tails, enterprise coding losses, and smaller
document/web harms---but are not a statistical loss sample.}
\label{tab:incident_checks}
\small
\setlength{\tabcolsep}{2pt}
\begin{tabular}{lp{0.47\linewidth}r}
\toprule
Category & Example & Reported loss \\
\midrule
Financial & Knight Capital deployment error in automated trading & \$440M \\
Financial & Air Canada chatbot fare-policy commitment & about \$1K \\
Coding & LangChain-style agent loop consuming compute without output & about \$47K \\
Coding & Coding agent deletion/outage reports & tens of thousands \\
Document & Mata v. Avianca hallucinated legal citations & \$5K sanction \\
Web & Autonomous shopping-bot purchase & \$31 \\
\bottomrule
\end{tabular}
\end{table}

These examples support only three weak claims: automated financial decisions
have the heaviest observed tail; coding-agent failures plausibly live in the
tens-of-thousands range for enterprise recovery and outage costs; and document
or web-agent losses often start smaller but can become material when multiplied
across users.  The evaluation therefore reports sensitivity results rather than
presenting the scenario table as a final rate filing.

\subsection{Additional Simulation and Ablations}
\label{app:portfolio}

We compare the CVaR-loaded trace premium against an ARS-style premium and
several formula ablations.  The relevant diagnostic results are:

Here ``solvency'' is the fraction of simulated one-period portfolio draws in
which collected premium covers realized claims and administrative cost; the
diagnostic credits neither outside capital nor reinsurance.

\begin{itemize}
\item An ARS-style collateral-discount premium is structurally underpriced in
our portfolio simulation: its maximum solvency is 4.5\% for
$\lambda\le 1.0$.
\item Replacing exogenous $\phat$ with trace-estimated $\phat$ improves
information but does not fix a defective premium formula; the tail term is
the critical ingredient.
\item At $\lambda=0.3$, CVaR-loaded pricing gives 23.8\% solvency versus
1.7\% for the ARS-style baseline, a 14$\times$ improvement.  At
$\lambda=1.0$, the gap is 79.8\% versus 4.5\%.
\item Removing the CVaR component collapses solvency to the low-single-digit
range even when the loading factor is increased.  This is why the main paper
keeps CVaR as a contract-design primitive.
\item No tested $(\kappa,\alpha_{\mathrm{CVaR}})$ pair reaches 95\% solvency
at $\lambda=0.3$; the best high-tail setting reaches roughly 43\% while
reducing adoption.  This is empirical support for reinsurance or policy caps,
not merely a tuning artifact.
\end{itemize}

These portfolio-solvency simulations are weaker evidence than the main
trace-economic evaluation because several loss labels are scenario-generated.
They remain valuable because the insolvency and tail-loading findings are
formula-level stress tests: they show what can go wrong even before one argues
about the best trace-risk predictor.
Table~\ref{tab:hyperparams_app} gives the simulation settings needed to
reproduce these formula-level stress tests.

\subsection{Simulation Hyperparameters}

\begin{table}[h]
\centering
\caption{\textbf{Portfolio stress-test settings.}  The simulation fixes episode
count, trace length, action-risk aggregation, premium loading, systemic-event
sweeps, and tail loading so that solvency results can be reproduced and
recalibrated.}
\label{tab:hyperparams_app}
\small
\setlength{\tabcolsep}{3pt}
\begin{tabular}{p{0.22\linewidth}p{0.42\linewidth}p{0.22\linewidth}}
\toprule
Parameter & Meaning & Value \\
\midrule
$N_{\mathrm{ep}}$ & episodes per run & 5,000 \\
Seeds & independent synthetic runs & 5 \\
$T$ & synthetic actions per session & 10 \\
$\kappa$ & mean/CVaR blend in $R(\tau)$ & 0.4 \\
$q$ & empirical CVaR tail level for action risk & 0.8 \\
$(a,b)$ & default sigmoid link & $(6.0,-3.0)$ \\
$\theta$ & proportional premium loading & 0.05 \\
$c_{\mathrm{admin}}$ & flat admin cost & \$50 \\
$\alpha_{\mathrm{CVaR}}$ & claim-loss CVaR level & 0.95 \\
$q_{\mathrm{sys}}$ & systemic event sweep & 0, 0.01, 0.05 \\
$\lambda$ & tail-loading sweep & 0--1.0 grid \\
\bottomrule
\end{tabular}
\end{table}

Synthetic sessions sample a latent task failure probability, perturb
action-level dimensions around that latent risk, aggregate by
$R(\tau)=(1-\kappa)\bar r+\kappa\cvar_q(r)$, and then estimate
$\phat=\sigma(6R-3)$.  This creates a controlled stress test for underwriting
policies.  It is not meant to validate trace-risk estimation by itself; the
real-trace SWE-smith and audit results serve that role.

\subsection{Per-Task Simulation Pattern}

Losses are also decomposed by task category.  We emphasize the qualitative
pattern because the exact synthetic values are less important than
the ordering.  Financial workflows dominate aggregate tail exposure because
severity is heavy-tailed and attribution matters for adjudication.  Coding
workflows produce moderate losses but strong control value because many
harmful actions are visible before deployment.  Document and web/support tasks
have lower per-episode severity but remain important at scale because they are
high frequency.  This pattern motivates the paper's recommendation: start with
sublimits for financial workflows, co-insurance for coding and web/support, and
lighter full coverage only for low-severity document tasks.

\subsection{Provider Break-Even and Contract Economics}

Provider NPV gives an economic interpretation of the viability frontier.  Let ACV be
average contract value and let $\Delta q$ be the incremental adoption caused by
offering coverage.  A safety commitment is NPV-positive when:
\begin{align*}
N\Delta q\,\mathrm{ACV} >
&N\E[\mathrm{Payout}(L)]\\
&+ \lambda\cvar_\alpha[L_{\mathrm{total}}]
+ c_{\mathrm{admin}}N .
\end{align*}
The required break-even adoption lift is therefore:
\[
\Delta q^\star =
\frac{c_{\mathrm{claim}}(C,D)(1-\rho_u)
+ \lambda\cvar_\alpha[L]/N + c_{\mathrm{admin}}}
{\mathrm{ACV}} .
\]
Tables~\ref{tab:breakeven_app} and~\ref{tab:contract_economics_app} report the
two quantities a provider pilot must estimate: whether coverage can pay for
itself through adoption lift, and which contract structure reduces claim cost
without destroying buyer value.

\begin{table}[h]
\centering
\caption{\textbf{Break-even adoption lift.}  Coverage is economically plausible
only when incremental buyer adoption exceeds expected claim, tail-capital, and
administrative costs; values above 35\% are treated as unlikely without a
mandate, pool, or backstop.}
\label{tab:breakeven_app}
\small
\setlength{\tabcolsep}{2pt}
\begin{tabular}{lccc}
\toprule
Provider profile & Full & 20\% co-ins. & Deduct. + sublimit \\
\midrule
Coding startup & 12.5\% & 10.4\% & 11.5\% \\
Financial provider & 42.8\% & 34.3\% & 19.8\% \\
General platform & 53.5\% & 43.0\% & 25.9\% \\
\bottomrule
\end{tabular}
\end{table}

\begin{table}[h]
\centering
\caption{\textbf{Contract economics.}  Coinsurance, deductibles, sublimits, and
exclusions reduce claim cost differently across workflows; the design question
is whether the risk reduction is large enough without eliminating adoption
value.}
\label{tab:contract_economics_app}
\small
\setlength{\tabcolsep}{2pt}
\begin{tabular}{p{0.30\linewidth}p{0.25\linewidth}cc}
\toprule
Structure & Recommended use & Claim cost & Adoption lift \\
\midrule
Full coverage & low-risk document/coding & baseline & baseline \\
20\% co-insurance & coding and web tasks & -20\% & small drop \\
\$2K deductible, \$500K cap & financial and tail-risk tasks & -53\% & small drop \\
Adversarial exclusion & all categories & -8\% & minimal drop \\
\bottomrule
\end{tabular}
\end{table}

The robust conclusion is ordinal, not the exact threshold: sublimits dominate
for heavy-tailed financial workflows; coinsurance is a low-friction way to
align incentives; full coverage is plausible only for lower-severity settings
or where a provider has unusually high measured adoption lift.

\subsection{Industry-Scale Liability Scenarios}

The social-impact stakes become visible at portfolio scale.  If a provider
covers $N$ customers with expected annual claim cost $m$, idiosyncratic
expected payouts are $Nm$.  A systemic model event adds approximately
$q_{\mathrm{sys}}N\E[L^{\mathrm{sys}}]$.  For example:

\begin{table}[h]
\centering
\caption{\textbf{Aggregate liability scale.}  Even modest per-customer expected
claim costs become large at platform scale, which is why systemic layers,
limits, and backstops are part of the insurance design rather than optional
deployment details.}
\label{tab:industry_scale}
\small
\setlength{\tabcolsep}{3pt}
\begin{tabular}{lccc}
\toprule
Scenario & $N$ & $m$ & $Nm$ \\
\midrule
Narrow enterprise pilot & 100K & \$1K & \$100M \\
Large SaaS assistant & 1M & \$10K & \$10B \\
Mass-market agent & 10M & \$5K & \$50B \\
\bottomrule
\end{tabular}
\end{table}

Table~\ref{tab:industry_scale} is only a scale calculation, but it explains why
systemic exposure cannot be treated as an implementation footnote.
With $q_{\mathrm{sys}}=1\%$ and $\E[L^{\mathrm{sys}}]=\$50K$, the common-shock
layer adds \$50M, \$500M, and \$5B respectively to the three rows.  These
numbers are illustrative, but they explain why the paper treats AI insurance
as social infrastructure rather than a narrow product feature: without
credible pricing, limits, auditability, and backstops, broad indemnity promises
can fail exactly when many harmed users need compensation simultaneously.

\subsection{Capacity Interpretation and Public Backstop}

These illustrative liabilities can be compared with existing cyber-insurance
capacity.  Exact market figures will change over
time, so the paper should avoid anchoring the main claim to a single current
premium-volume estimate.  The structural interpretation is more durable:

\begin{itemize}
\item idiosyncratic agent failures can be priced and pooled privately when
claims are bounded and trace-verifiable;
\item systemic model-update shocks are correlated across all customers of the
same provider and therefore do not diversify inside that provider's portfolio;
\item heavy-tailed financial or medical workflows require policy limits,
catastrophic reinsurance, mandatory pools, or public backstops;
\item an AI insurance regime should therefore separate a private layer for
ordinary agent-caused harms from a public or pooled layer for systemic shocks.
\end{itemize}

This preserves the analogy to flood, terrorism, and cyber markets without
making the paper depend on a fragile point estimate of industry capacity.

\subsection{Sequential Trace-Risk Validation}

The main paper reports the compact AUC comparison because the central claim is
economic rather than purely predictive.  We also test whether trace-risk
features become informative early enough to support live intervention as
secondary evidence for the control-policy design.

\begin{figure*}[t]
\centering
\includegraphics[width=0.90\textwidth]{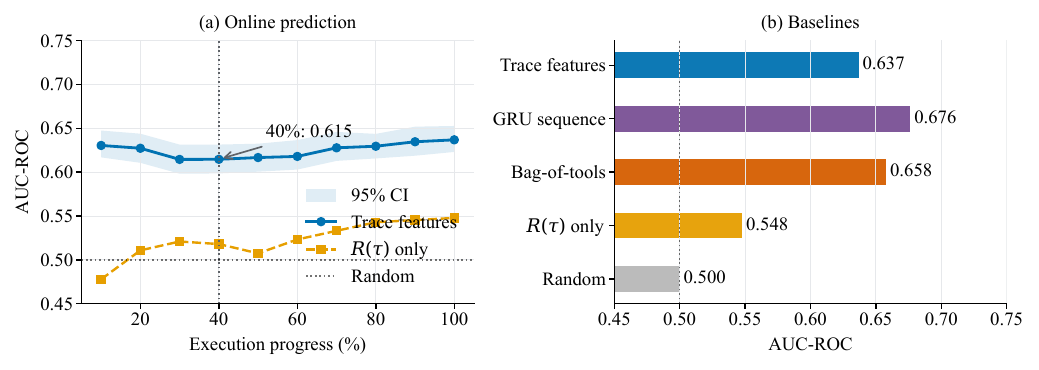}
\caption{\textbf{Sequential monitoring.}  Interpretable trace features become
predictive before the end of a trajectory and recover much of the trained
sequence baseline's above-random signal; the figure supports review triggers,
not calibrated insurance prices.}
\label{fig:sequential_auc_app}
\end{figure*}

The validation treats SWE-smith task failure as a proxy label for behavioral
risk, not as a direct insurance claim.  Features are computed from the trace
without using the outcome label; only the two scalar link parameters in
$\phat=\sigma(aR(\tau)+b)$ are calibrated.  On 5,000 SWE-smith trajectories,
the interpretable model reaches AUC 0.637, compared with 0.676 for a GRU over
tool sequences.  Both can run online, but the trace score is auditable and
recovers 78\% of the GRU's improvement above random while remaining usable for
underwriting.
Table~\ref{tab:swebench_auc_app} and Figure~\ref{fig:sequential_auc_app}
therefore support live monitoring, not final actuarial calibration.

\begin{table}[h]
\centering
\caption{\textbf{Auditable sequence signal.}  The trace-risk model is not the
highest-AUC predictor, but it is live and interpretable; it recovers much of
the sequence-model ranking signal while fitting only scalar calibration
parameters.}
\label{tab:swebench_auc_app}
\small
\setlength{\tabcolsep}{3pt}
\begin{tabular}{lcccc}
\toprule
Model & AUC & Training & Live & Interpretable \\
\midrule
GRU on tool sequence & 0.676 & yes & yes & no \\
Bag-of-tools LR & 0.658 & yes & yes & partial \\
Trace-risk model & 0.637 & scalar only & yes & yes \\
$R(\tau)$ only & 0.548 & no & yes & yes \\
Mean action risk only & 0.539 & no & yes & yes \\
Random & 0.500 & -- & -- & -- \\
\bottomrule
\end{tabular}
\end{table}

Two caveats keep this result from being overclaimed.  First, task failure is
only a proxy for claimable harm.  It is useful because many insurable harms
require task failure, but it is not a substitute for adjudicated claims.
Second, a tau-bench smoke test showed that the sign of the link can reverse in
policy-constrained customer-service tasks: low-action
traces may fail because the agent is blocked from acting.  This is why the
paper treats $(a,b)$ calibration as domain-specific rather than assuming a
universal scalar direction.  Sequence baselines such as the GRU also depend on
clean training data, and entropy-based filtering of poisoned trajectories
\citep{xu2025clip} would be a prerequisite before treating any learned baseline
as a production underwriting signal.

\subsection{Risk-Dimension Ablation}
\label{app:dim_ablation}

To assess how much each of the five risk dimensions contributes to the trace-priced
risk score $\hat{p}$, we run a leave-one-dimension-out ablation on a held-out pool of
$n=5{,}000$ episodes drawn from the evaluation's task-profile mixture (30 seeds).

\textbf{Metrics.}
We report two complementary metrics:
(1)~\emph{Spearman $\rho$}, the rank correlation of $\hat{p}$ with $p_{\text{true}}$,
which is scale-invariant and therefore insensitive to OLS calibration; and
(2)~\emph{pricing MAE}, the mean absolute error of the OLS trace-priced premium
against $\hat{L}=p_{\text{true}}\times\mathrm{sev\_mean}$.

\begin{figure}[h]
\centering
\includegraphics[width=0.92\columnwidth]{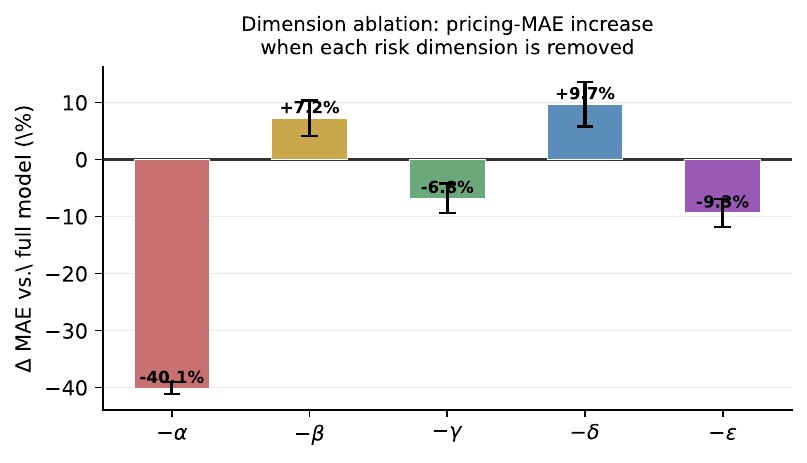}
\caption{\textbf{Dimension-ablation effect.}  Removing temporal position or
blast radius hurts linear premium calibration, while the anomalous
irreversibility result is an OLS linearisation artefact; the paired
Table~\ref{tab:dim_ablation} shows that irreversibility remains the dominant
rank-ordering signal.}
\label{fig:dim_ablation}
\end{figure}

\textbf{Results} are shown in Table~\ref{tab:dim_ablation}
(see also Figure~\ref{fig:dim_ablation}).

\begin{table}[t]
\centering
\caption{\textbf{Risk-dimension ablation.}  Leave-one-dimension-out results
($n=5{,}000$ training episodes, 30 seeds) show which trace fields carry
ordering signal and which mainly affect linear premium calibration.
\textit{Spearman $\rho$}: rank correlation of $\hat{p}$ with $p_{\mathrm{true}}$
(scale-invariant, primary metric).
\textit{$\Delta\rho$}: change relative to full model; a negative $\Delta\rho$ means
the dimension is informative for risk ordering.
$\alpha$ shows the largest rank-information drop ($-$20.8\%), confirming it is the
dominant signal; its apparent MAE improvement under OLS is a linearisation artefact
(the multiplicative gate $r_t=\alpha_t\!\cdot\!\mathrm{inner}$ creates a
$p_{\mathrm{true}}^2$ relationship that OLS cannot exploit).}
\label{tab:dim_ablation}
\small
\setlength{\tabcolsep}{4pt}
\begin{tabular}{lccc}
\toprule
Model & Spearman $\rho$ & $\Delta\rho$ & $\Delta$ MAE \\
\midrule
\textbf{Full model}             & \textbf{0.948} & ---          & --- \\
$-\alpha$ (irreversibility)     & 0.751          & $-$20.8\%    & $-$40.1\%\textsuperscript{\dag} \\
$-\beta$ (blast radius)         & 0.942          & $-$0.7\%     & $+$7.2\% \\
$-\gamma$ (epistemic)           & 0.943          & $-$0.5\%     & $-$6.8\% \\
$-\delta$ (temporal)            & 0.948          & $\approx$0   & $+$9.7\% \\
$-\varepsilon$ (causal)         & 0.955          & $+$0.8\%     & $-$9.3\% \\
\bottomrule
\multicolumn{4}{l}{\textsuperscript{\dag}OLS artefact; see caption.}
\end{tabular}
\end{table}

\textbf{Interpretation.}
Spearman $\rho$ reveals that $\alpha$ (irreversibility) is by far the most
informative dimension for \emph{ranking} risky episodes: removing it cuts the rank
correlation from $0.948$ to $0.751$ ($-$20.8\%).
This is consistent with the public incident checks in
Appendix~\ref{app:calibration}: every documented
loss~\cite{wolak2025,pocketos2026,chiu2026,aiid1152} was triggered by an
irreversible write---\texttt{rm -rf}, \texttt{volumeDelete}, \texttt{DROP TABLE}---
for which $\alpha_t=1.0$.

The $\alpha$ dimension \emph{gates} the composite risk
multiplicatively ($r_t = \alpha_t \cdot \mathrm{inner}_t$).
This is semantically correct: a high-blast-radius action that is nonetheless
reversible (e.g.\ a staging-environment write) carries less systemic risk than a
low-blast-radius irreversible delete.  However, the multiplicative structure
creates a $p_{\text{true}}^2$-like relationship with $\hat{p}$, which OLS
cannot exploit, explaining the apparent $-40.1\%$ MAE ``improvement'' when $\alpha$
is ablated.  The rank-correlation metric is not affected by this nonlinearity.

$\delta$ (temporal position) contributes a calibration shift rather than an
ordinal signal: its Spearman contribution is $\approx 0$ but its MAE contribution
is $+9.7\%$, because the temporal ramp affects the overall magnitude of $R(\tau)$
and hence the OLS intercept.  $\beta$ (blast radius) and $\gamma$ (epistemic) each
provide small but real rank contributions ($-0.7\%$ and $-0.5\%$ Spearman).
$\varepsilon$ (causal attribution) is slightly redundant with $\alpha$ in simulation;
in real traces it may carry independent signal from multi-agent provenance chains.

The takeaway for practitioners: a simplified scorer that retains
$\alpha$ and $\beta$ captures most of the ranking signal; $\gamma$, $\delta$, $\varepsilon$
matter for premium calibration more than for episode triage.

\textbf{Extreme weight robustness.}
To assess robustness beyond the bounded perturbations in Table~\ref{tab:dim_ablation},
we test a qualitatively different weighting scheme: a minimal model using only
$\alpha$ (irreversibility gate) and $\beta$ (blast radius), with $\gamma$, $\delta$, $\varepsilon$
set to zero.
This $(\alpha,\beta)$-only model achieves Spearman $\rho=0.921$ (vs $0.948$ full)
on the 30-seed ablation.  It therefore retains most of the full model's ranking
signal under an extreme simplification, confirming that the key ordering is not
an artefact of the specific five-dimensional weighting.  We do not report its
absolute dollar MAE here because the ablation is a feature diagnostic; the
primary pricing MAE is the 5-seed portfolio result reported above.

\subsection{Adoption-Lift Model and Viability Sensitivity}

The main text uses the viability frontier to motivate contract structure.  A
compact adoption-lift model identifies which business quantities must be
measured before a provider can responsibly promise coverage.

\begin{figure}[h]
\centering
\includegraphics[width=0.98\linewidth]{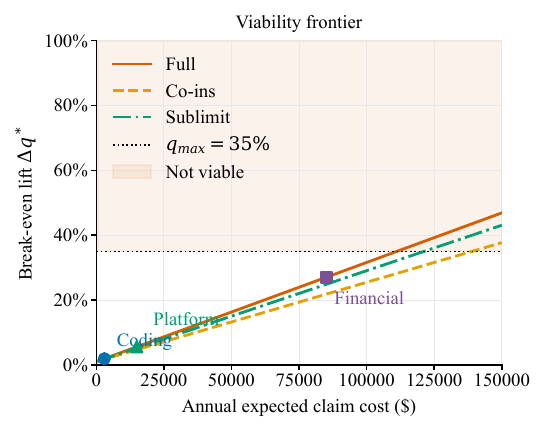}
\caption{\textbf{Viability frontier.}  Full coverage is difficult for
heavy-tailed workflows; sublimits and coinsurance move the commitment into a
more plausible private-market region.}
\label{fig:viability_frontier_app}
\end{figure}

\begin{figure}[h]
\centering
\includegraphics[width=0.98\linewidth]{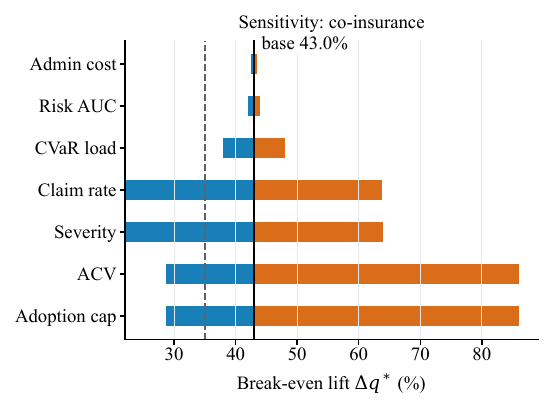}
\caption{\textbf{Adoption-lift sensitivity.}  The dominant uncertainty is not a
small predictive-model gain; it is whether the commitment measurably changes
buyer behavior and how heavy the claim severity tail is.}
\label{fig:tornado_app}
\end{figure}

Let $g(C,D,\rho_u)\in[0,1]$ denote coverage generosity.  The adoption lift from
a safety commitment is modeled as
\[
\Delta q(C,D,\rho_u)=q_{\max}\left(1-e^{-\phi g(C,D,\rho_u)}\right).
\]
Here $q_{\max}=0.35$ is a scenario ceiling for the share
of enterprise buyers whose adoption is materially affected by liability
coverage, and $\phi=0.56$ was an illustrative response parameter.  These values
should not be read as measured demand for agent insurance.  Their value is
diagnostic: they let the paper ask whether a claimed safety commitment would
need implausibly large adoption lift to break even.
Table~\ref{tab:claim_cost_app} and
Figures~\ref{fig:viability_frontier_app}--\ref{fig:tornado_app} decompose this
question into claim cost, viable contract region, and sensitivity drivers.

\begin{table}[h]
\centering
\caption{\textbf{Claim-cost ordering.}  Financial workflows dominate expected
claim cost, while document and support tasks are lower-severity; these scenario
values explain the viability-frontier ordering, not final rates.}
\label{tab:claim_cost_app}
\small
\setlength{\tabcolsep}{3pt}
\begin{tabular}{lcccc}
\toprule
Task & Full & 20\% co-ins. & Deduct. + sublimit & $\phat_{\rm avg}$ \\
\midrule
Financial & \$75.2K & \$60.2K & \$33.8K & 15\% \\
Coding & \$2.9K & \$2.3K & \$2.7K & 12\% \\
Document & \$0.3K & \$0.2K & \$0.2K & 7\% \\
Web/support & \$1.3K & \$1.1K & \$1.1K & 10\% \\
Mixed & \$19.9K & \$15.9K & \$9.4K & 11\% \\
\bottomrule
\end{tabular}
\end{table}

The practical lesson is that a provider pilot should measure adoption
lift and willingness-to-pay directly, because those quantities dominate the
viability conclusion.  Better trace scoring improves segmentation and control,
but it cannot make a full-coverage commitment viable when expected claim cost
and tail capital exceed the economic value of additional adoption.

\subsection{Dynamic Premium Update and Alert Rule}

The evaluation treats interventions as policy decisions; the same trace signal
can also operate inside a monitored contract.  The update rule clarifies that monitoring
is contractual and auditable rather than an ad hoc post-hoc charge.

At session start, before any action is observed, the policy uses a task-domain
prior $\phat_0=\sigma(b)$ and computes an initial premium $P_0$ from the
CVaR-loaded premium formula.  As the trace unfolds, the running risk estimate is
\[
\phat_t=\sigma(aR(\tau_{1:t})+b),
\]
and the dynamic premium index is
\[
P_t=P_0\frac{\phat_t}{\phat_0}.
\]
The ratio can move down as well as up: a session consisting only of reversible
read actions can become cheaper, while an irreversible production write or
financial operation increases the risk index.

The control rule used in the case study is:
\[
\mathrm{Alert}_t = 1
\quad\Longleftrightarrow\quad
\phat_t>\phat_{\max}\ \mathrm{and}\ \alpha_t>\alpha_{\min}.
\]
The irreversibility gate is important.  It prevents alert fatigue when the
aggregate session looks risky but the current action is still read-only.  A
practical deployment would also include a pre-commitment clause: if the user
halts and restarts immediately before a high-risk action, billing and review
should use the alert-time risk index rather than allowing strategic avoidance.
Computationally, the update is lightweight; a running top-$k$ structure makes
the empirical CVaR update effectively constant time per action for typical
agent traces.

\subsection{Portfolio Formula Diagnostics}

\begin{figure*}[t]
\centering
\includegraphics[width=0.90\textwidth]{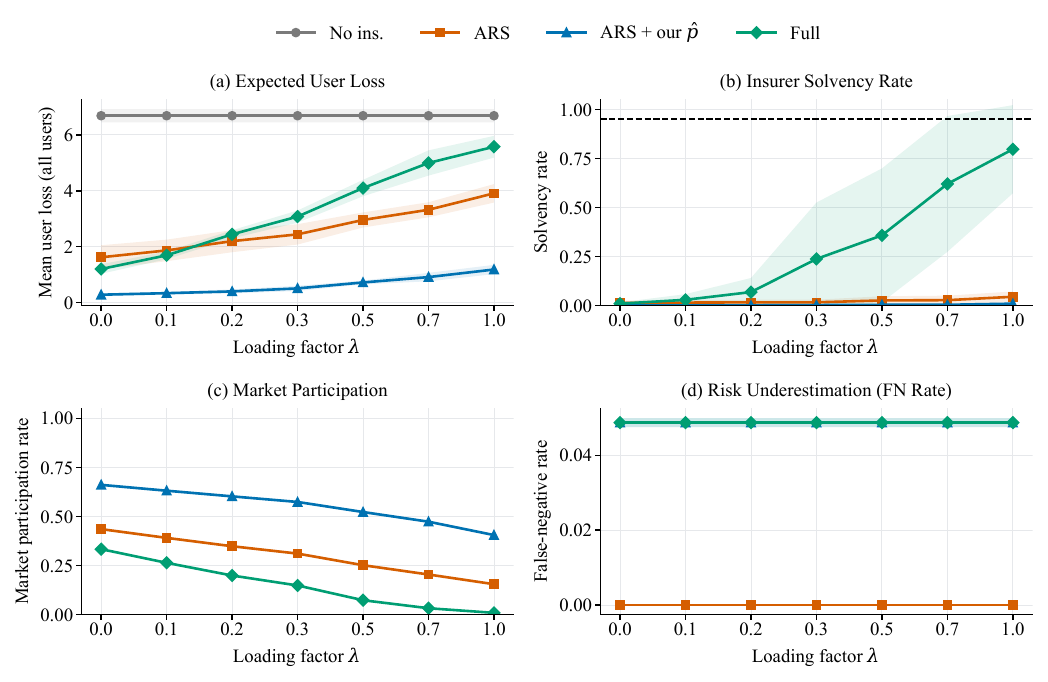}
\caption{\textbf{Loading-factor sweep.}  Raising tail loading improves solvency
but lowers adoption; the diagnostic shows why the insurance mechanism must
report both insurer solvency and buyer participation.}
\label{fig:legacy_main_app}
\end{figure*}

\begin{figure*}[t]
\centering
\includegraphics[width=0.90\textwidth]{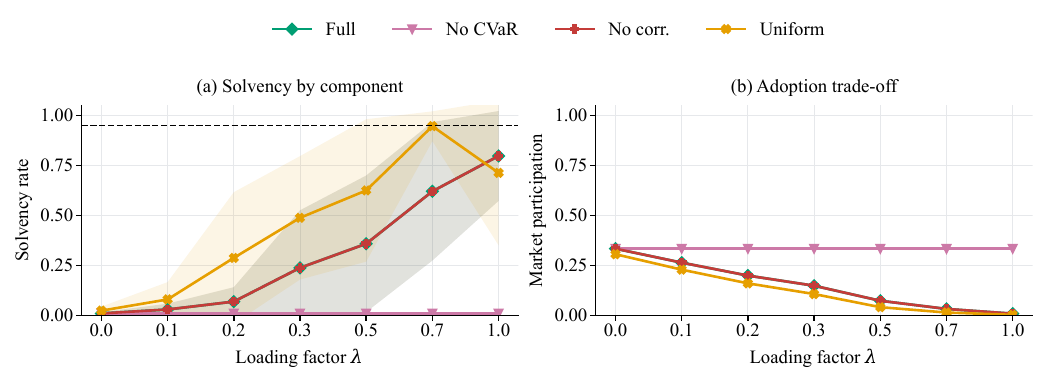}
\caption{\textbf{Tail-capital ablation.}  Removing the CVaR component collapses
solvency even when expected-loss information is available, supporting tail
capital as a primitive in the contract model.}
\label{fig:legacy_ablation_app}
\end{figure*}

\begin{figure*}[t]
\centering
\includegraphics[width=0.90\textwidth]{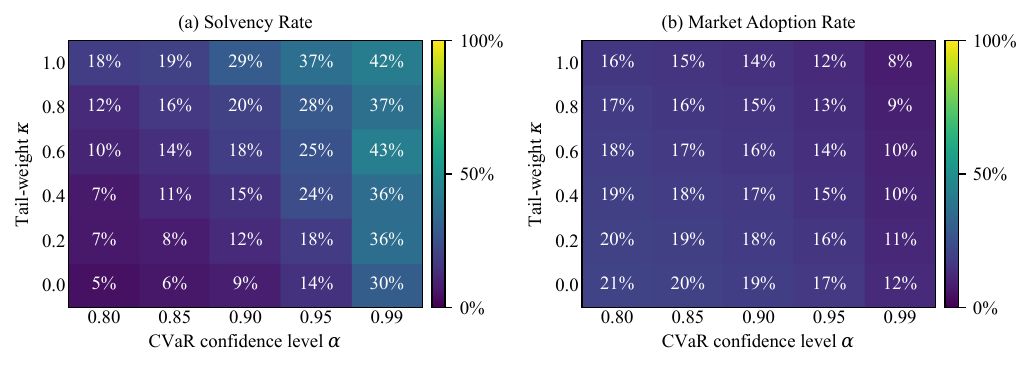}
\caption{\textbf{Hyperparameter stress.}  Higher tail sensitivity improves
solvency but lowers adoption, and no tested setting reaches a conventional
solvency target at low loading; this motivates limits, reinsurance, or
backstops rather than more tuning.}
\label{fig:legacy_heatmap_app}
\end{figure*}

\begin{figure*}[t]
\centering
\includegraphics[width=0.90\textwidth]{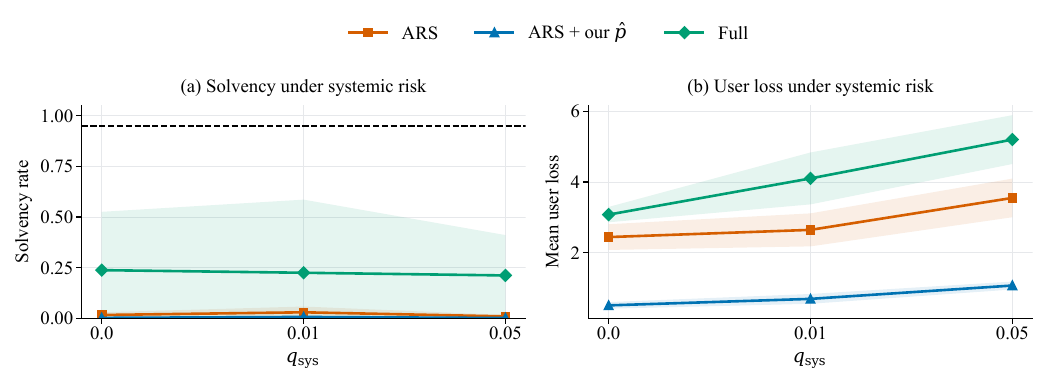}
\caption{\textbf{Systemic-risk sweep.}  In the tested low-prevalence range,
systemic events are second-order relative to premium-formula underpricing, but
the analytical common-shock condition still requires caps, reinsurance, or a
backstop for high-stakes deployments.}
\label{fig:legacy_systemic_app}
\end{figure*}

We also evaluate an ARS-style synthetic portfolio.  This simulation is not the primary evidence in
the final paper because its risk scores are synthetic and correlated with
ground truth by construction.  It is still useful as a formula-level diagnostic:
it shows that tail loading and solvency constraints matter even if the risk
probability were known.  Table~\ref{tab:legacy_portfolio_app} and
Figures~\ref{fig:legacy_main_app},~\ref{fig:legacy_ablation_app},
~\ref{fig:legacy_heatmap_app}, and~\ref{fig:legacy_systemic_app} therefore
support one narrow claim: better risk information does not rescue a premium
formula that omits tail capital and common-shock design.

\begin{table}[h]
\centering
\caption{\textbf{Formula diagnostic at $\lambda=0.3$.}  Synthetic portfolio
results isolate premium-formula behavior: ARS-style pricing can attract buyers
while remaining insolvent, whereas CVaR loading improves solvency by explicitly
charging for tail loss.}
\label{tab:legacy_portfolio_app}
\small
\setlength{\tabcolsep}{1.5pt}
\begin{tabular}{lcccc}
\toprule
Condition & Mean loss & CVaR$_{95}$ & Solv. & Adopt. \\
\midrule
No insurance & 6.69 & 29.96 & -- & 0.0\% \\
ARS escrow-only & 2.44 & 0.00 & 1.7\% & 31.1\% \\
ARS + trace $\phat$ & 0.51 & 0.00 & 0.2\% & 57.4\% \\
CVaR-loaded & 3.08 & 10.42 & 23.8\% & 14.8\% \\
\bottomrule
\end{tabular}
\end{table}

The scientific value of this simulation is deliberately narrow: trace-derived $\phat$ is not
enough.  A liability mechanism also needs a premium formula that prices tails
and a market design that handles common shocks.

\subsection{Contract Clauses}
\label{app:contracts}

Three clauses are especially important for agentic AI.

\textbf{Adversarial input exclusion.}
Coverage should not apply when a third party causes harm through prompt
injection, unauthorized parameter manipulation, or transfer attacks from
surrogate models \citep{xu2025one}, unless the insured followed specified
input-validation controls and the provider contract assumes that residual
risk.

\textbf{Model-version continuity.}
Coverage should specify whether losses under a new model version are covered.
Major model upgrades can materially change failure rates and should trigger
re-underwriting or a revised premium.

\textbf{Deployment-context warranty.}
Users should disclose task category, autonomy level, transaction limits, and
production exposure.  Moving a general coding agent into medical, financial, or
legal decision-making without endorsement changes the risk class and should
void or modify coverage.

\textbf{Regulatory-compliance condition.}
Coverage should be conditional on required conformity assessments, human
oversight obligations, and domain-specific AI rules being satisfied at the time
of deployment.  This matters economically because unlawful deployment changes
both expected loss and claim defensibility.

\subsection{Trace Verifiability Score}

For a loss event $\ell$ and trace $\tau$, define an information-based
verifiability score
\[
v_{\mathrm{MI}}(\ell,\tau)=\frac{I(\ell\mid\tau)}{H(\ell)},
\]
where $I(\ell\mid\tau)=H(\ell)-H(\ell\mid\tau)$ is the reduction in uncertainty
about the cause of the loss after observing the trace.  In practice this can be
approximated by counterfactual replay: if replacing a particular action with a
null action sharply reduces the estimated loss probability, and the action has
high attribution score $\epsilon_t$, the claim has high trace verifiability.
This claim-time information score extends, but is not identical to, the
testbed's action-level proxy $v=\operatorname{mean}(\epsilon_{1:T})$.
This score motivates the three-class taxonomy used in the paper:

\begin{itemize}
\item high verifiability: logs directly show the harmful write, deletion, API
call, or transaction;
\item partial verifiability: expert analysis is needed to link agent advice or
task completion to downstream loss;
\item low verifiability: prompt injection, compound human-agent decisions, or
undisclosed deployment context make causation unreliable.
\end{itemize}

When raw logs are too coarse to attribute a harmful action, VLM-based
diagnosis of the trace artifacts \citep{xu2026internal} can supplement
counterfactual replay, although both methods remain auxiliary to the
deterministic attribution score.

\subsection{Additional Limitations}

Two limitations remain important for future versions.
First, epistemic uncertainty $\gamma_t$ is only weakly measured in the current
trace conversion because many public traces are single-run logs.  Coarse action
type self-consistency is not enough: agents may agree on the tool class while
disagreeing on the specific file, query, recipient, or transaction parameter.
Fine-grained tool-call self-consistency would make $\gamma_t$ more informative.
Second, the framework addresses insurable losses: quantifiable, verifiable, and
bounded harms.  Catastrophic or existential AI risks need separate governance
and public-risk frameworks rather than ordinary indemnity contracts.

\end{document}